\documentclass[10pt,twocolumn,letterpaper]{article}
\usepackage[pagenumbers]{cvpr} 

\usepackage[accsupp]{axessibility}
\usepackage[dvipsnames]{xcolor}
\usepackage{amsmath,amssymb,amsfonts}
\usepackage{algorithmic}
\usepackage{graphicx}
\usepackage{textcomp}
\usepackage{xcolor}
\usepackage{color}
\usepackage{colortbl}
\usepackage{mathtools}
\usepackage{amsmath}
\usepackage{tabularx}
\usepackage{multirow}
\usepackage{multicol}
\usepackage{float}
\usepackage{tabularx}
\usepackage{threeparttable}
\usepackage{booktabs}
\usepackage{longtable}
\usepackage{makecell}
\usepackage{verbatim}
\usepackage{wrapfig}
\usepackage{xspace}
\usepackage{pbox}
\usepackage{epstopdf}
\usepackage{comment}
\usepackage{lipsum}
\usepackage{bm}
\usepackage{bbm}
\usepackage{setspace}
\usepackage{sidecap}
\usepackage{array}
\usepackage{blindtext}
\usepackage{enumitem}
\usepackage{diagbox}
\usepackage{bbding}
\usepackage{url}
\usepackage{indentfirst}
\definecolor{cvprblue}{rgb}{0.21,0.49,0.74}
\usepackage[pagebackref,breaklinks,colorlinks,citecolor=cvprblue]{hyperref}

\usepackage{svg}

\usepackage[capitalize]{cleveref}
\crefname{section}{Sec.}{Secs.}
\crefname{section}{Section}{Sections}
\crefname{table}{Table}{Tables}
\crefname{table}{Tab.}{Tabs.}

\newcommand{\dataset}{CORE4D\xspace}

\def\paperID{} 
\def\confName{CVPR}
\def\confYear{2025}

\title{CORE4D: A 4D Human-Object-Human Interaction Dataset for Collaborative Object REarrangement}

\author{
Yun Liu\textsuperscript{*,1,2,3},~
Chengwen Zhang\textsuperscript{*,1,4},~
Ruofan Xing\textsuperscript{1},~
Bingda Tang\textsuperscript{1},~
Bowen Yang\textsuperscript{1},~
Li Yi\textsuperscript{\textdagger,1,2,3}
\smallskip\\
\textsuperscript{1}Tsinghua University~~~
\textsuperscript{2}Shanghai Qi Zhi Institute~~~
\textsuperscript{3}Shanghai Artificial Intelligence Laboratory\\
\textsuperscript{4}Beijing University of Posts and Telecommunications\\
\url{https://core4d.github.io/}
}

\begin{document}
\twocolumn[{
\renewcommand\twocolumn[1][]{#1}
\maketitle
\begin{center}
    \captionsetup{type=figure}
    \includegraphics[width=0.95\linewidth]{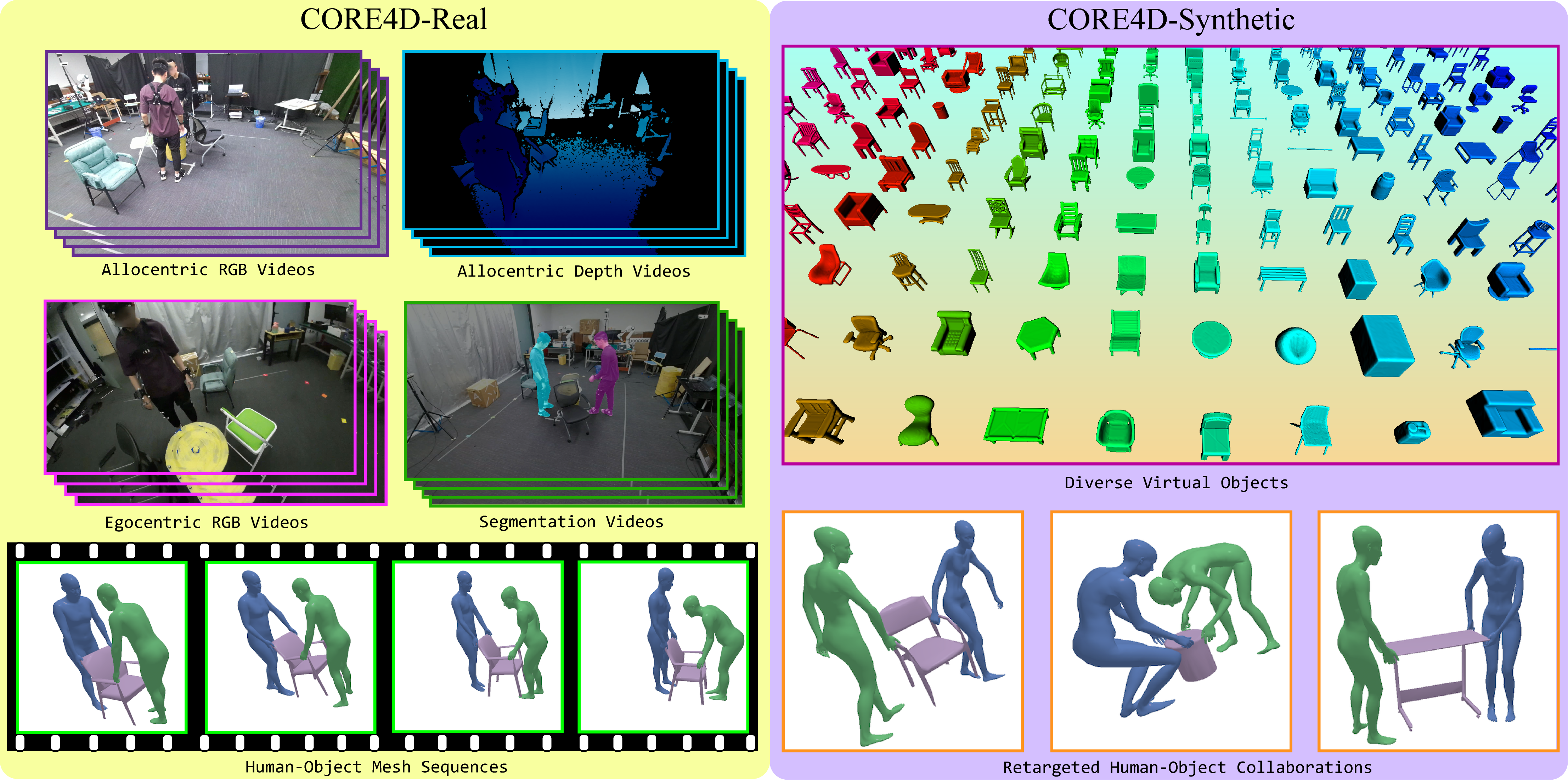}
    \caption{\dataset is a large-scale diverse human-object-human interaction dataset for collaborative object rearrangement, encompassing real-world and synthetic branches. CORE4D-Real captures 1K human-object-human mesh sequences with allocentric and egocentric visual signals, while CORE4D-Synthetic retargets real-world data onto 3K virtual object shapes formulating 10K motion sequences.}
    \label{fig:teaser}
\end{center}
}]

\footnotetext[1]{Equal contribution.}
\footnotetext[2]{Corresponding author.}

\begin{abstract}
Understanding how humans cooperatively rearrange household objects is critical for VR/AR and human-robot interaction. However, in-depth studies on modeling these behaviors are under-researched due to the lack of relevant datasets. We fill this gap by presenting \dataset, a novel large-scale 4D human-object-human interaction dataset focusing on collaborative object rearrangement, which encompasses diverse compositions of various object geometries, collaboration modes, and 3D scenes. With 1K human-object-human motion sequences captured in the real world, we enrich \dataset by contributing an iterative collaboration retargeting strategy to augment motions to a variety of novel objects. Leveraging this approach, \dataset comprises a total of 11K collaboration sequences spanning 3K real and virtual object shapes. Benefiting from extensive motion patterns provided by \dataset, we benchmark two tasks aiming at generating human-object interaction: human-object motion forecasting and interaction synthesis. Extensive experiments demonstrate the effectiveness of our collaboration retargeting strategy and indicate that \dataset has posed new challenges to existing human-object interaction generation methodologies.

\label{sec:abstract}
\end{abstract}

\section{Introduction}
\label{sec:introduction}
Humans frequently rearrange household items through multi-person collaboration
, such as moving a table or picking up an overturned chair together. Analyzing and synthesizing these diverse collaborative behaviors could be widely applicable in VR/AR, human-robot interaction~\cite{wan2022learn,ng2023takes,ng2023diffusion}, dexterous manipulation~\cite{christen2022d,xu2023unidexgrasp,wan2023unidexgrasp++,zhang2024artigrasp} and humanoid manipulation~\cite{murooka2021humanoid,dao2023sim,xie2023hierarchical,li2023kinodynamics}. However, understanding and modeling these interactive motions have been limited due to the lack of large-scale, richly annotated datasets. Most existing human-object and hand-object interaction datasets focus on individual behaviors~\cite{GRAB,BEHAVE,HOI4D,yang2022oakink,ARCTIC,CHAIRS,NeuralDome,OMOMO,TACO,OAKINK2} and two-person handovers~\cite{ye2021h2o,HOH,InteractiveHumanoid}. But these datasets typically encompass a limited number of object instances, thus struggling to support generalizable interaction understanding across diverse object shapes.
Scaling up precise human-object interaction data is challenging. While vision-based human-object motion tracking methods~\cite{xie2023visibility,zhao2024m,xie2024template,xie2024intertrack} have advanced, they still struggle with low fidelity due to severe occlusions, which are common in multi-human collaborations. Additionally, mocap~\cite{CHAIRS,OMOMO} is expensive and hard to scale up to cover numerous objects involved in rearrangement. Our goal is to curate a large-scale category-level human-object-human (HOH) interaction dataset with high quality in a cost-efficient manner.

We observe that HOH collaborations mainly vary in two aspects: the temporal collaboration patterns of two humans and the spatial relations between human and object. The temporal collaboration patterns could vary widely depending on scene complexity, motion range, and collaboration mode. In contrast, the spatial relations between human and object tend to possess strong homogeneity when facing objects from the same category, e.g., two persons holding opposite sides of a chair.  
This allows for retargeting interactions involving one specific instance to another using automatic algorithms, avoiding the need to capture interactions with thousands of same-category objects in the real world.
The above observations enable us to leverage expensive motion capture systems to capture only humans' diverse temporal collaboration patterns, while relying on automatic spatial retargeting algorithms to enrich human-object spatial relations.

Using these insights, we build a novel large-scale dataset, \dataset, encompassing a wide range of human-object interactions for collaborative object rearrangement. \dataset includes various types of household objects, collaboration modes,
and 3D environments. Our data acquisition strategy combines mocap-based capturing and synthetic retargeting, allowing us to scale the dataset effectively. The retargeting algorithm transfers spatial relation between human and object to novel object geometries while preserving temporal pattern of human collaboration. As a result, \dataset includes 1K real-world motion sequences (\dataset-Real) paired with videos and 3D scenes, as well as 10K synthetic collaboration sequences (\dataset-Synthetic) covering 3K diverse object shapes.


We benchmark two tasks for generating human-object collaboration: (1) motion forecasting~\cite{CAHMP,InterDiff} and (2) interaction synthesis~\cite{MDM,OMOMO} on \dataset, revealing challenges in modeling human behaviors, enhancing motion naturalness, and adapting to new object geometries. 
Ablation studies demonstrate the effectiveness of our hybrid data acquisition strategy, and the quality and value of \dataset-Synthetic, highlighting its role in helping to improve existing motion generation methods. We further retarget interactions in CORE4D onto Unitree H1~\cite{unitree_h1} humanoid robot and use them to train humanoid box-lifting policies, showcasing the values of \dataset in robot interaction skill learning.




In summary, our main contributions are threefold: (1) We present \dataset, a large-scale 4D HOH interaction dataset for collaborative object rearrangement. (2) We propose a novel hybrid data acquisition method, incorporating real-world data capture and synthetic collaboration retargeting. (3) We benchmark two tasks for collaboration generation, revealing new challenges and research opportunities.



\section{Related Work}
\subsection{Human-object Interaction Datasets}


Tremendous progress has been made in constructing human-object interaction datasets. 
To study how humans interact with 3D scenes, various widely-used datasets record human movements and surrounding scenes separately, treating objects as static~\cite{cao2020long,savva2016pigraphs,PROX,GPA,SAMP,4DCapture,HSC4D,EgoBody,COUCH,HPS,HUMANISE,CIRCLE,RICH,CIMI4D,Sloper4D,guzov2022interaction,tanke2024humans,yan2024reli11d,jiang2024scaling,yi2024generating,jiang2024autonomous} or partially deformable~\cite{MocapDeform} without pose changes. 
For dynamic objects, recent works~\cite{carfi2019multi,cini2019choice,chan2020affordance,khanna2023multimodal,kshirsagar2023dataset,thoduka2024multimodal,GRAB,GraviCap,BEHAVE,InterCap,CHAIRS,NeuralDome,OMOMO,HOH,InteractiveHumanoid,FORCE,HOI-M3,P3HAOI,zhao2024m,lv2025himo,F-HOI} have captured human-object interaction behaviors with varying focuses. 
Table \ref{tab:dataset_comparison} generally summarizes the characteristics of 4D human-object-interaction datasets.
To support research for vision-based human-object motion tracking and shape reconstruction, a line of datasets~\cite{GraviCap,BEHAVE,InterCap,CHAIRS,NeuralDome,FORCE,HOI-M3,P3HAOI} present human-object mesh annotations with multi-view RGB or RGBD signals. 
With the rapid development of human-robot cooperation, several works~\cite{carfi2019multi,GRAB,HOH,InteractiveHumanoid} focus on specific action types such as grasping~\cite{GRAB} and human-human handover~\cite{carfi2019multi,HOH,InteractiveHumanoid}.
Our dataset uniquely captures multi-person and object collaborative motions, category-level interactions, and both egocentric and allocentric views, offering comprehensive features with the inclusion of both real and synthetic datasets.

\begin{table*}[tb]
  \centering
  \footnotesize
  \begin{tabular}{|c|ccccccccc|}
    \hline
    \multirow{2}{*}{dataset} & multi- & \multirow{2}{*}{collaboration} & category- & \multirow{2}{*}{egocentric} & \multirow{2}{*}{RGBD} & \multirow{2}{*}{$\#$view} & \multirow{2}{*}{mocap} & \multirow{2}{*}{$\#$object} & \multirow{2}{*}{$\#$sequence} \\
    & human & & level & & & & & & \\
    \hline
    GRAB~\cite{GRAB} & & & & & & - & \checkmark & 57 & - \\
    GraviCap~\cite{GraviCap} & & & & & & 3 & & 4 & 9 \\
    BEHAVE~\cite{BEHAVE} & & & & & \checkmark & 4 & & 20 & 321 \\
    InterCap~\cite{InterCap} & & & & & \checkmark & 6 & & 10 & 223 \\
    CHAIRS~\cite{CHAIRS} & & & \checkmark & & \checkmark & 4 & \checkmark & 81 & 1.4K \\
    HODome~\cite{NeuralDome} & & & & & & 76 & \checkmark & 23 & 274 \\
    \textit{Li et.al.}~\cite{OMOMO} & & & & & & - & \checkmark & 15 & 6.1K \\
    FORCE~\cite{FORCE} & & & & & \checkmark & 1 & \checkmark & 8 & 450 \\
    IMHD$^2$~\cite{zhao2024m} & & & & & & 32 & \checkmark & 10 & 295 \\
    HIMO~\cite{lv2025himo} & & & & & & - & \checkmark & 53 & 3.4K \\
    \textit{Carfi et.al.}~\cite{carfi2019multi} & \checkmark & & & & \checkmark & 1 & \checkmark & 10 & 1.1K \\
    HOH~\cite{HOH} & \checkmark & & & & \checkmark & 8 & & 136 & 2.7K \\
    CoChair~\cite{InteractiveHumanoid} & \checkmark & & \checkmark & & & - & \checkmark & 8 & 3.0K \\
    HOI-M$^3$~\cite{HOI-M3} & \checkmark & & & & & 42 & \checkmark & 90 & 199 \\
    \hline
    \textbf{\dataset-Real} & \checkmark & \checkmark & \checkmark & \checkmark & \checkmark & 5 & \checkmark & 37 & 1.0K \\
    \textbf{\dataset-Synthetic} & \checkmark & \checkmark & \checkmark & & & - & - & \textbf{3.0K} & \textbf{10K} \\
  \hline
  \end{tabular}
  \vspace{-0.2cm}
  \caption{\textbf{Comparison of \dataset with existing 4D human-object interaction datasets.}}
  \vspace{-0.5cm}
  \label{tab:dataset_comparison}
\end{table*}

\subsection{Human Interaction Retargeting}

Human interaction retargeting focuses on applying human interactive motions to novel objects in human-object interaction scenarios. Existing methodologies~\cite{kim2016retargeting,rodriguez2018transferring,yang2022oakink,simeonov2022neural,wu2023functional,ISAGrasp,huang2024spatial,xie2024template} are object-centric. They propose first finding contact correspondences between the source and the target objects and then adjusting human motion to touch specific regions on the target object via optimization. As crucial guidance for the result, contact correspondences are discovered by aligning either surface regions~\cite{rodriguez2018transferring,yang2022oakink,wu2023functional,xie2024template}, spatial maps~\cite{kim2016retargeting,huang2024spatial}, distance fields~\cite{ISAGrasp}, or neural descriptor fields~\cite{simeonov2022neural} between the source and the target objects. These methods are all limited to objects with similar topology and scales. Our synthetic data generation strategy incorporates object-centric design~\cite{yang2022oakink} with novel human-centric contact selection, allowing adaptation to challenging objects using human priors.

\subsection{Human-object Interaction Generation}


Human-object interaction generation is an emerging research topic that aims to synthesize realistic human-object motions conditioned on surrounding 3D scenes, known object trajectories, or action types. 
To generate 3D human mesh snapshots interacting with scenes, POSA~\cite{POSA} and COINS~\cite{COINS} propose to leverage CVAE~\cite{CVAE}, 
while DreamHOI~\cite{zhu2024dreamhoi} provides an iterative NeRF~\cite{mildenhall2021nerf} optimization approach. To further synthesize interactive motions, a line of works~\cite{NSM,IMOS,SAGA,GOAL,COUCH,NIFTY,ROAM,mir2024generating} present auto-regressive manners~\cite{NSM,COUCH}, diffusion models~\cite{NIFTY}, or two-stage designs that first generates start and end poses and then interpolates motion in-between~\cite{IMOS,SAGA,GOAL,ROAM}. Beyond static objects, a line of works further model object movements and generate integrated human-object interactions using diffusion models~\cite{InterDiff,OMOMO} and GCN~\cite{yan2024forecasting}.
To generate human-object interactions under action descriptions, recent works~\cite{CG-HOI,CHOIS,HOI-Diff,wu2024thor,InterDreamer,wu2024human,wang2024move,yi2025generating,fan2024textim,jiang2024autonomous,song2024hoianimator} extract text features with pretrained CLIP encoders~\cite{radford2021learning,wang2024move,yi2025generating,jiang2024autonomous,song2024hoianimator} or LLM planners~\cite{ChatGPT,touvron2023llama,wu2024human,fan2024textim}, using them to guide diffusion models~\cite{ho2020denoising}.

\section{Constructing CORE4D}

\dataset is a large-scale 4D human-object-human interaction dataset acquired in a novel hybrid scheme, comprising \dataset-Real and \dataset-Synthetic. \dataset-Real is captured (Section \ref{sec:data_capturing}) and annotated (Section \ref{sec:data_annotating}) from authentic collaborative scenarios. As shown in Figure~\ref{fig:dataset}, it provides human-object-human poses, egocentric RGB videos, allocentric RGB-D videos, and 2D segmentations across 1.0K sequences accompanied by 37 object models. 
To augment spacial relation between human and object,
we present an innovative collaboration retargeting technique in Section \ref{sec:retargeting}. This technique integrates \dataset-Real with \dataset-Synthetic, thereby expanding our collection with an additional 10K sequences and 3K rigid objects. Detailed characteristics such as data diversities are discussed in Section \ref{sec:dataset_characteristics}.

\subsection{\dataset-Real Data Capture}
\label{sec:data_capturing}
\begin{figure}[h]
\vspace{-0.5cm}
  \centering
  \includegraphics[width=1.0\linewidth]{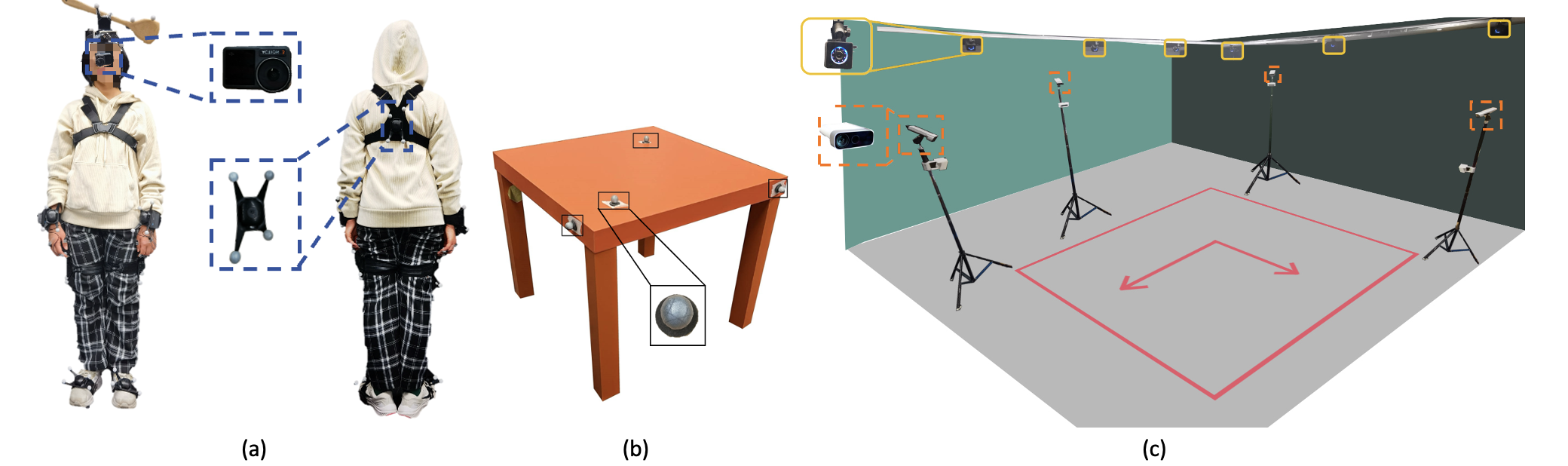}
  \caption{\textbf{CORE4D-Real data capturing system.} (a) demonstrates the wearing of mocap suits and the positioning of the egocentric camera. (b) shows an object with four markers. (c) illustrates the data capturing system and camera views.}
  \label{fig:data_capturing_system}
\vspace{-0.3cm}
\end{figure}

\begin{figure*}[ht]
  \centering
  \includegraphics[width=1.0\textwidth]{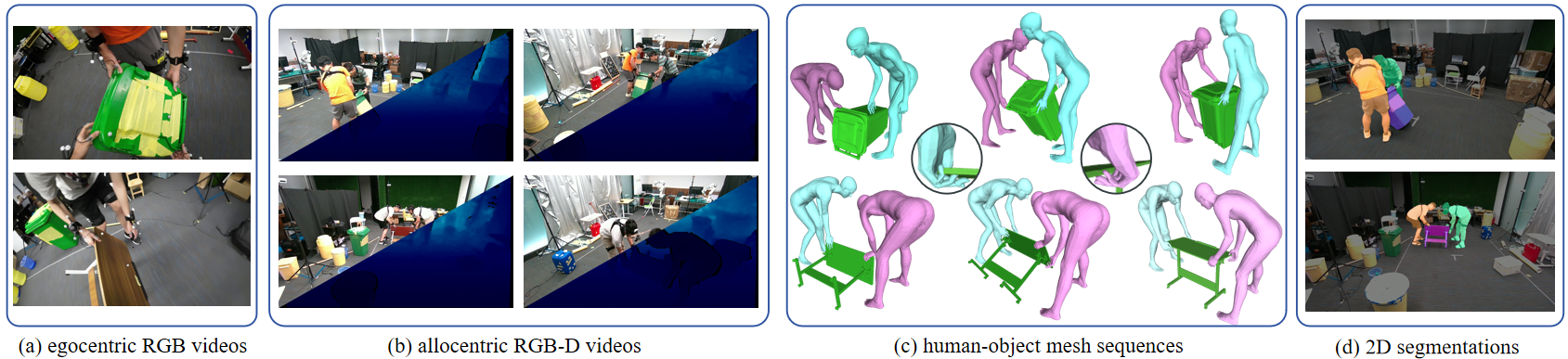}
  \vspace{-0.7cm}
  \caption{\textbf{CORE4D-Real data modality overview.}}
  \vspace{-0.3cm}
  \label{fig:dataset}
\end{figure*}



To collect precise human-object motions with visual signals, we set up a hybrid data capturing system shown in Fig. \ref{fig:data_capturing_system}, consisting of an inertial-optical mocap system, four allocentric RGB-D cameras and a camera worn by participants for egocentric sensing. The system operates at 15 FPS.

\noindent\textbf{Inertial-optical Mocap System.} To accurately capture human-object poses in multi-person collaboration scenarios, often involving severe occlusion, we use an inertial-optical mocap system~\cite{noitom} inspired by CHAIRS~\cite{CHAIRS}
This system includes 12 infrared cameras, mocap suits with 8 inertial-optical trackers and two data gloves per person, and markers of a 10mm radius. The mocap suits capture Biovision Hierarchy (BVH) skeletons of humans, while markers attached to the objects track object motion.



\noindent\textbf{Visual Sensors.}
Kinect Azure DK cameras are integrated to capture allocentric RGB-D signals, and an Osmo Action3 is utilized to capture egocentric color videos. The resolution of all the visual signals is 1920x1080. Cameras are calibrated by the mocap system and synchronized via timestamp. Details on camera calibration and synchronization are in the appendix.

\noindent\textbf{Object Model Acquisition.} \dataset-Real includes 37 3D models of rigid objects spanning six household object categories. Each object model is constructed by an industrial 3D scanner with up to 100K triangular faces. We additionally adopt manual refinements on captured object models to remove triangle outliers and improve accuracy.

\noindent\textbf{Privacy Protection.} To ensure participant anonymity, blurring is applied to faces~\cite{haarcascades} in RGB videos. The participants all consented to releasing \dataset, and were notified of their right to remove their data from \dataset at any time.

\subsection{\dataset-Real Data Annotation}
\label{sec:data_annotating}

\noindent\textbf{Object Pose Tracking.} To acquire the 6D pose of a rigid object, we attach four to five markers to the object's surface. The markers formulate a virtual rigid that the mocap system can track. With accurate localization of the object manually, the object pose can be precisely determined by marker positions captured by the infrared cameras.

\noindent\textbf{Human Mesh Acquisition.} Aligning with existing datasets~\cite{CHAIRS,OMOMO}, we retarget BVH~\cite{meredith2001motion} human skeletons to SMPL-X~\cite{SMPLX}. SMPL-X~\cite{SMPLX} formulates a human mesh as
$D_{\text{smplx}} = M(\beta, \theta)$. The body shape $\beta \in \mathbb{R}^{10}$ are optimized to fit the constraints on manually measured human skeleton lengths.
With $\beta$ computed, we optimize the full-body pose $\theta \in \mathbb{R}^{159}$ with the loss function:
\vspace{-0.1cm}
\begin{align}
    \mathcal{L} = \mathcal{L}_{\text{reg}} + \mathcal{L}_{j\text{3D}} + \mathcal{L}_{j\text{Ori}} + \mathcal{L}_{\text{smooth}} + \mathcal{L}_{h\text{3D}} + \mathcal{L}_{h\text{Ori}} + \mathcal{L}_{\text{contact}},
\end{align}

\vspace{-0.2cm}
\noindent where $\mathcal{L}_{\text{reg}}$ ensures the simplicity of the results and prevents unnatural, significant twisting of the joints. $\mathcal{L}_{j\text{3D}}$ and $\mathcal{L}_{j\text{Ori}}$ encourage the rotation of joints and the global 3D positions to closely match the ground truth. $\mathcal{L}_{h\text{3D}}$ and $\mathcal{L}_{h\text{Ori}}$ guide the positioning and orientation of the fingers. $\mathcal{L}_{\text{smooth}}$ promotes temporal smoothness. $\mathcal{L}_{\text{contact}}$ encourages realistic contact between the hands and objects. Then using SMPL-X~\cite{SMPLX} $M(\beta, \theta, \Phi) : \mathbb R^{|\theta| \times |\beta|} \mapsto \mathbb R^{3N}$ to generate human mesh. Details on loss functions are in the appendix.


\noindent\textbf{2D Mask Annotation.} We offer automatic 2D segmentation for individuals and the manipulated objects to aid in predictive tasks like vision-based human-object pose estimation~\cite{BEHAVE,xie2023visibility}. We first use DEVA~\cite{cheng2023tracking} to segment human and object instances in a captured interaction image with text prompts. Then, we render human and object meshes separately on each image and select the instance with the highest Intersection-over-Union (IoU) for mask annotation.




\begin{figure*}[h]
  \centering
  \includegraphics[width=1.0\textwidth]{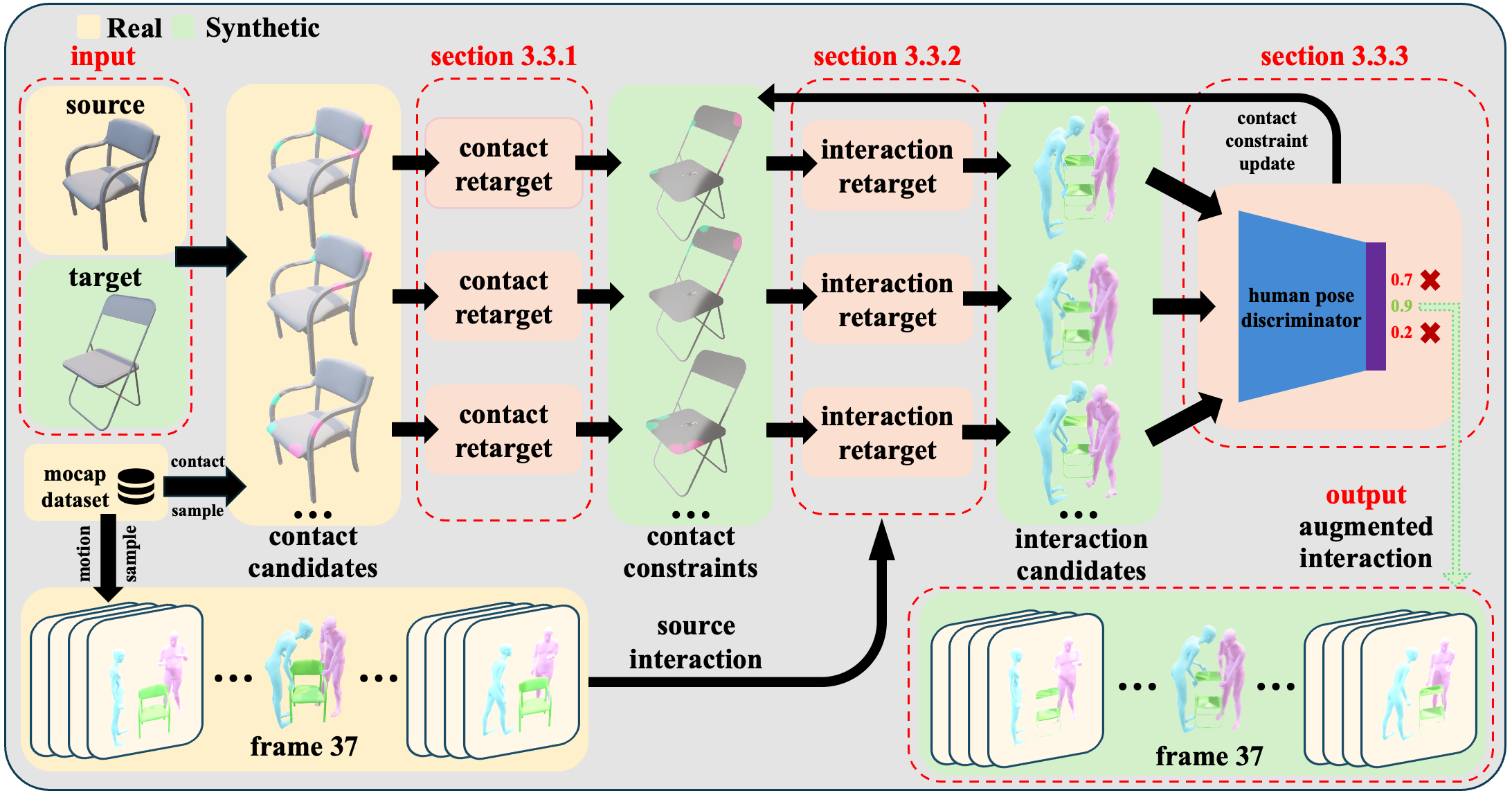}
  \caption{\textbf{Collaboration retargeting pipeline.} We propose a collaboration retargeting algorithm by iteratively refining interaction motion. The input is a \textit{source}-\textit{target} pair. First, we sample contact candidates from whole \dataset-Real contact knowledge on \textit{source}. For each contact candidate, we apply contact retargeting to propagate contact candidates to contact constraints on \textit{target}. Sampled motion from \dataset-Real provides a high-level collaboration pattern, together with low-level contact constraints, we obtain interaction candidates from interaction retargeting. Then, the human pose discriminator selects the optimal candidates, prompting a contact constraints update via beam search.
  After multiple iterations, the process yields augmented interactions. This iterative mechanism can effectively get a reasonable one from numerous contact constraints and ensures a refined interaction, enhancing the dataset's applicability across various scenarios.}
  \vspace{-0.3cm}
  \label{fig:retargeting_pipeline}
\end{figure*}

\subsection{\dataset-Synthetic Data Generation}
\label{sec:retargeting}


In order to enrich the diversities of object geometries and human-object spatial relations,
our retargeting algorithm transfers real interactions to ShapeNet~\cite{chang2015shapenet} objects of the same category, thereby significantly expanding the dataset regarding the object's diversity. 
When transferring interactions across objects, contact points are always the key and it is important to consider whether they can be properly transferred with consistent semantics on new objects~\cite{yang2021cpf,zheng2023cams}. However, we find this insufficient when object geometries vary largely and correspondences become hard to build. We thus tackle interaction retargeting from a novel human-centric perspective where good contact points should support natural human poses and motions. 
We realize this idea through the pipeline depicted in Figure \ref{fig:retargeting_pipeline},
which comprises three key components.
First, \textbf{object-centric contact retargeting} uses whole contact knowledge from \dataset-Real to obtain accurate contact with different objects. 
Second, \textbf{contact-guided interaction retargeting} adapts motion sequences to new object geometries while considering the contact constraints.
Third, a \textbf{human-centric contact selection} evaluates poses from interaction candidates to select the most plausible contacts.

\noindent\textbf{Object-centric Contact Retargeting.}
\label{sec:contact_retargeting}
To acquire reasonable human poses, contact constraints on the target object are essential. We draw inspiration from Tink~\cite{yang2022oakink} and train DeepSDF on all objects' signed distance fields (SDFs). For \textit{source} object SDF $O_{s}$ and \textit{target} object SDF $O_{t}$, we first apply linear interpolation on their latent vectors $o_{s}$ and $o_{t}$ and obtain $N$ intermediate vectors $o_i = \frac{N+1-i}{N+1} o_s + \frac{i}{N+1} o_t (1\leq i\leq N)$. We then decode $o_i$ to its SDF $O_i$ via the decoder of DeepSDF, and reconstruct the corresponding 3D mesh $M_i$ using the Marching Cubes algorithm~\cite{lorensen1998marching}.
Thereby get mesh sequence $\mathcal{M} = [\textit{source}, M_1, M_2, ..., M_N, \textit{target}]$ and successively transfer contact positions between every two adjacent meshes in $\mathcal{M}$ via Nearest-neighbor searching. In addition, we leverage all contact candidates from \dataset-Real on \textit{source} to form a pool of contact candidates and transfer them to \textit{target} as contact constraints. 





\noindent\textbf{Contact-guided Interaction Retargeting.}
\label{sec:CIR}
For each contact constraint, 
interaction retargeting aims to transfer human interaction from \textit{source} to \textit{target}. To greatly enforce the consistency of interaction motion, we 
optimize
variables including
the object rotations $R_o \in \mathbb{R}^{N\times3}$ and translations $T_o \in \mathbb{R}^{N\times3}$, human poses $\theta_{1,2} \in \mathbb{R}^{2 \times N \times 153}$, translation $T_{1,2}\in \mathbb{R}^{2 \times N \times 3}$ and orientation $O_{1,2}\in \mathbb{R}^{2 \times N \times 3}$ on the SMPL-X~\cite{SMPLX}. $N$ is the frame number. 
We first estimate the \textit{target}'s motion $\{R_o,T_o\}$ by solving an optimization problem as follows:
\vspace{-0.1cm}
\begin{align}
    R_o, T_o \longleftarrow \mathop{\operatorname{argmin}}\limits_{R_o, T_o}(\mathcal{L}_{f} + \mathcal{L}_{\text{spat}} + \mathcal{L}_{\text{smooth}}),
\end{align}
where fidelity loss $\mathcal{L}_f$ evaluates the difference of the \textit{target}'s rotation and translation against the \textit{source}, restriction loss $\mathcal{L}_{\text{spat}}$ penalizes \textit{target}'s penetration with the ground, and smoothness loss $\mathcal{L}_{\text{smooth}}$ constrains the \textit{target}'s velocities between consecutive frames.

Given the \textit{target}'s motion and contact constraints, we then transfer humans' interactive motion $\{\theta_{1,2},T_{1,2},O_{1,2}\}$ from the \textit{source} to the \textit{target} by solving another optimization problem as follows:
\begin{align}
    \theta_{1,2}, T_{1,2}, O_{1,2} \longleftarrow \mathop{\operatorname{argmin}}\limits_{\theta_{1,2}, T_{1,2}, O_{1,2}}(\mathcal{L}_{j} + \mathcal{L}_{c} + \mathcal{L}_{\text{spat}} + \mathcal{L}_{\text{smooth}}),
\end{align}
where fidelity loss $\mathcal{L}_{\text{j}}$ evaluates the difference in human joint positions before and after the transfer, contact loss $\mathcal{L}_{c}$ computes the difference between human-object contact regions and the contact constraints, $\mathcal{L}_{\text{spat}}$ and $\mathcal{L}_{\text{smooth}}$ ensures the smoothness of human motion. Details on the loss designs are in the appendix.

\noindent\textbf{Human-centric Contact Selection.}
\label{sec:human_pose_discriminator}
Selecting reasonable contact constraints efficiently is challenging due to their large scales and the time-consuming interaction retargeting.
We address this challenge by developing a beam search algorithm to select contact constraints from a human-centric perspective. Specifically, we train a human pose discriminator inspired by GAN-based motion generation works~\cite{yan2019convolutional,xu2023actformer}.
To train it, we build a pairwise training dataset, with each pair consisting of one positive human pose sample and one negative one. Positive samples are encouraged to get higher scores than negative ones. 
We use \dataset-Real as positive samples.
We add 6D pose noise $\Delta(\alpha, \beta, \gamma, x, y, z)$ on \textit{target} motion, and regard corresponding human motions generated by contact-guided interaction retargeting as negative samples.
The loss function is:
\begin{align}
\mathcal{L}_{\text{ranking}} = - \log(\sigma(R_{\text{pos}} - R_{\text{neg}} - m(S_{\text{pos}}, S_{\text{neg}}))),
\end{align}
\noindent where $S_{\text{pos}}$ and $S_{\text{neg}}$ denote inputs for positive and negative samples respectively, with $R_{\text{pos}}$ and $R_{\text{neg}}$ being their corresponding discriminator scores. $\sigma$ is Sigmoid function, and $m(S_{\text{pos}}, S_{\text{neg}}) = ||\Delta(\alpha, \beta, \gamma, x, y, z)||$ is human-guide margin~\cite{RLHF} between positive and negative poses. This margin could explicitly instruct the discriminator to yield more significant disparities across different poses.

To ensure the realism of human interactions, we also introduce an interpenetration penalty.
We prioritize those with the highest discriminator scores while ensuring acceptable levels of interpenetration as the optimal contact constraints. 

\subsection{Dataset Characteristics}
\label{sec:dataset_characteristics}

\begin{figure*}[h]
  \centering
  \includegraphics[width=\textwidth]{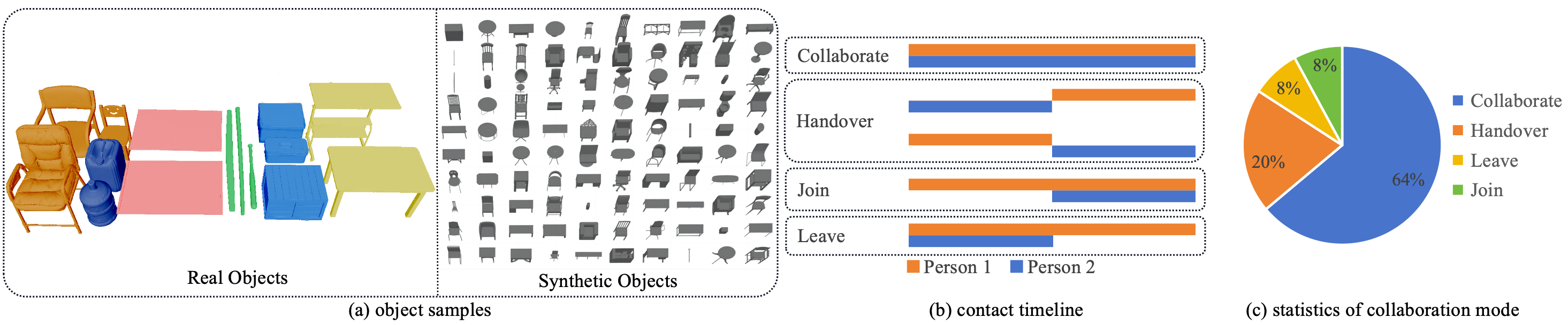}
  \vspace{-0.4cm}
  \caption{\textbf{Dataset statistics.} (a) shows object samples from six categories. Bars in (b) indicate when the person is in contact with the object during the entire collaborative object rearrangement interaction process. (c) presents the proportion of collaboration modes in the dataset.}
  \vspace{-0.3cm}
  \label{fig:dataset_statistics}
\end{figure*}

To better model collaborative object rearrangement interactions, we focus on diversifying our dataset in several vital areas: object geometries, collaboration modes, and 3D scenes. These ensure a comprehensive representation of real-world interactions.

\noindent\textbf{Diversity in Object Geometries.} We design six object categories to cover the main collaborative object rearrangement interaction scenarios as Fig. \ref{fig:dataset_statistics}(a). Categories with relatively simple geometry, uniformity, and typically exhibiting symmetry include box, board, barrel, and stick. Categories with more complex geometries and significant individual differences include chair and desk.

\noindent\textbf{Diversity in Collaboration Modes.} We define five human-human collaboration modes in collaborative object rearrangement. Each mode represents a unique form of collaboration between two individuals, providing a new perspective and possibilities for understanding and researching collaborative behaviors. 
At first, we define the person with the egocentric camera as Person 2, and the other as Person 1. Collaborative carrying tasks are divided by whether Person 2 knows the goal or not. Tasks of handover and solely move alternate between the two participants. In join and leave tasks, Person 2 will either join in to help or leave halfway through, respectively.

\noindent\textbf{Diversity in 3D Scenes.} Surrounding scenarios are set up with varying levels of scene complexity: no obstacle, single obstacle, and many obstacles (more than one). Participants are asked to navigate through these randomly placed obstacles by their own means. We observe that this typically involved behaviors including bypassing, going through, stepping over, or moving obstacles aside.

\section{Experiments}
In this section, we first present the train-test split of \dataset (Section \ref{sec:data_split}). We then propose two benchmarks for generating human-object collaboration: human-object motion forecasting (Section \ref{sec:motion_forecasting}), and interaction synthesis (Section \ref{sec:interaction_synthesis}). Finally, Section \ref{sec:evaluate_retargeting} presents extensive studies on the collaboration retargeting approach.

\subsection{Data Split}
\label{sec:data_split}


We construct a training set from a random assortment of real objects, combining their real motions and corresponding synthetic data.
We also create two test sets from \dataset-Real for non-generalization and inner-category generalization studies. Test set S1 includes interactions with training set objects, while S2 features interactions with new objects. \dataset-Synthetic is not included in the test set, avoiding potential biases from the retargeting algorithm.
Details are shown in the appendix.




\subsection{Human-object Motion Forecasting}
\label{sec:motion_forecasting}

Forecasting 4D human motion~\cite{guo2022multi,peng2022somoformer,mao2022contact,guo2023back} is a crucial problem with applications in VR/AR and embodied perception~\cite{kasahara2017malleable}. Current research~\cite{corona2020context,adeli2021tripod,wan2022learn,InterDiff} is limited to individual behaviors due to data constraints. Our work expands this by using diverse multi-person collaborations, making the prediction problem both intriguing and challenging.


\noindent\textbf{Task Formulation.} 
Given the object's 3D model and human-object poses in adjacent 15 frames, the task is to predict their subsequent poses in the following 15 frames. The human pose $P_h \in \mathbb{R}^{23 \times 3}$ represents joint rotations of the SMPL-X~\cite{SMPLX} model, while the object pose $P_o = \{R_o \in \mathbb{R}^3, T_o \in \mathbb{R}^3\}$ denotes 3D orientation and 3D translation of the rigid object model.


\noindent\textbf{Evaluation Metrics.} Following existing motion forecasting works~\cite{CAHMP,HO-GCN,InterDiff}, we evaluate human joints position error $J_e$, object translation error $T_e$, object rotation error $R_e$, human-object contact accuracy $C_{\text{acc}}$, and penetration rate $P_r$.
Details are provided in the appendix.






\noindent\textbf{Methods, Results, and Analysis.} We evaluate three state-of-the-art motion forecasting methods, MDM~\cite{MDM}, InterDiff~\cite{InterDiff}, and CAHMP~\cite{CAHMP}.
Table \ref{tab:results_motion_forecasting} 
 quantitatively shows these methods reveal a consistent drop in performance for unseen objects (S2) versus seen ones (S1) regarding human pose prediction. Meanwhile, errors in object pose prediction remain similar. This highlights the challenges in generalizing human collaborative motion for novel object shapes.

\begin{table}[tb]
  \centering
  \scriptsize
  \begin{tabular}{|c|c|c|c|c|c|c|}
    \hline
    Test & \multirow{2}{*}{Method} & Human & \multicolumn{2}{c|}{Object} & \multicolumn{2}{c|}{Contact} \\
    \cline{3-7}
    Set & & $J_e$ ($\downarrow$) & $T_e$ ($\downarrow$) & $R_e$ ($\downarrow$) & $C_{\text{acc}}$ ($\uparrow$) & $P_r$ ($\downarrow$) \\
    \hline
    \multirow{3}{*}{S1} & MDM~\cite{MDM} & 170.8 & 136.8 & 10.7 & 84.9 & 0.3 \\
    \cline{2-7}
    & InterDiff~\cite{InterDiff} & 170.8 & 135.1 & 10.2 & 84.9 & 0.3 \\
    \cline{2-7}
    & CAHMP~\cite{CAHMP} & 169.4 & 110.3 & 9.0 & - & - \\
    
    \hline
    \multirow{3}{*}{S2} & MDM~\cite{MDM} & 186.4 & 136.0 & 11.1 & 88.0 & 0.3 \\
    \cline{2-7}
    & InterDiff~\cite{InterDiff} & 186.4 & 133.6 & 10.7 & 88.0 & 0.3 \\
    \cline{2-7}
    & CAHMP~\cite{CAHMP} & 170.5 & 112.9 & 9.5 & - & - \\
  \hline
  \end{tabular}
  \vspace{-0.2cm}
  \caption{\textbf{Quantitative results on motion forecasting.}}
  \vspace{-0.4cm}
  \label{tab:results_motion_forecasting}
\end{table}

\subsection{Interaction Synthesis}
\label{sec:interaction_synthesis}

Generating human-object interaction~\cite{OMOMO,CG-HOI,CHOIS,HOI-Diff} is an emerging research topic benefiting human avatar animation and human-robot collaboration~\cite{christen2023learning,ng2023takes}. With extensive collaboration modes and various object categories, \dataset constitutes a knowledge base for studying generalizable algorithms of human-object-human interactive motion synthesis.

\noindent\textbf{Task Formulation.} Following recent studies~\cite{GOAL,OMOMO}, we define the task as object-conditioned human motion generation. Given an object geometry sequence $G_o \in \mathbb{R}^{T \times N \times 3}$, the aim is to generate corresponding two-person collaboration motions $M_h \in \mathbb{R}^{2 \times T \times 23 \times 3}$. This involves frame numbers $T$, object point clouds $G_o$, and human pose parameters for the SMPL-X~\cite{SMPLX} model.



\noindent\textbf{Evaluation Metrics.} Following individual human-object interaction synthesis~\cite{OMOMO}, we evaluate human joint position error $R.J_e$, object vertex position error $R.V_e$, and human-object contact accuracy $C_{\text{acc}}$.
The FID score (\textit{FID}) is leveraged to quantitatively assess the naturalness of synthesized results. Details of the metric designs are presented in the appendix.

\noindent\textbf{Methods, Results, and Analysis.} We utilize three advanced diffusion models~\cite{MDM,OMOMO,CHOIS} as baselines. MDM~\cite{MDM} and CHOIS~\cite{CHOIS} are one-stage conditional motion diffusion models, while OMOMO is a two-stage approach with hand positions as intermediate results. Quantitative evaluations reveal larger errors in OMOMO when modeling multi-human collaboration compared to individual interaction synthesis by \textit{Li et al.}~\cite{OMOMO}. Furthermore, the synthesized results have a higher FID than real motion data, indicating challenges in motion naturalness.

\begin{table}[tb]
  \centering
  \scriptsize
  \begin{tabular}{|c|c|c|c|c|c|}
    \hline
    Test Set & Method & $R.J_e$ ($\downarrow$) & $R.V_e$ ($\downarrow$) & $C_{\text{acc}}$ ($\uparrow$) & \textit{FID} ($\downarrow$) \\
    \hline
    \multirow{3}{*}{S1} & MDM~\cite{MDM} & 138.3 & 194.8 & 76.5 & 7.5 \\
    \cline{2-6}
    & OMOMO~\cite{OMOMO} & 138.0 & 196.9 & 78.0 & 7.8 \\
    \cline{2-6}
    & CHOIS~\cite{CHOIS} & 138.4 & 194.3 & 76.2 & 7.7  \\
    \hline
    \multirow{3}{*}{S2} & MDM~\cite{MDM} & 146.1 & 208.3 & 76.6 & 7.9 \\
    \cline{2-6}
    & OMOMO~\cite{OMOMO} & 145.3 & 209.9 & 77.8 & 7.4 \\
    \cline{2-6}
    & CHOIS~\cite{CHOIS} & 145.8 & 206.7 & 76.2 & 7.7 \\
    \hline
  \end{tabular}
  \vspace{-0.2cm}
  \caption{\textbf{Quantitative results on interaction synthesis.}}
  \vspace{-0.4cm}
  \label{tab:results_synthesis}
\end{table}

\subsection{Collaboration Retargeting}
\label{sec:evaluate_retargeting}

\begin{table}[ht]
    \centering
    \scriptsize
    \begin{tabular}{|cc|ccc|c|c|c|c|}
      \hline
      \multicolumn{2}{|c|}{\multirow{3}{*}{Comparisons}}       & \multicolumn{3}{c|}{Designs}      &    \multicolumn{2}{c|}{Phys. Eval.}       &\multicolumn{2}{c|}{User Preferences}      \\
      \cline{3-6} \cline{7-9}
      & &  \multirow{2}{*}{C} & \multirow{2}{*}{D} & \multirow{2}{*}{U} & $P$ & $C_{\text{acc}}$  & Contact & Motion    \\
      \cline{8-9}
         &         &       &        &         &($\downarrow$)  & ($\uparrow$)   &\multicolumn{2}{c|}{A / B / Approx. ($\uparrow$)}\\
      \hline
       \multirow{2}{*}{Abl.1}  &A  & &&       & 0.61 & 83.2  & \multirow{2}{*}{7.8/\textbf{88.9}/3.3}&\multirow{2}{*}{3.3/\textbf{84.4}/12.3}\\
         & B  &   \checkmark          & \checkmark         &          &  \textbf{0.24}        & \textbf{83.3}         &          &           \\
      \hline
      \multirow{2}{*}{Abl.2}  &A  & \checkmark && & 0.55 & 82.9   &\multirow{2}{*}{1.2/\textbf{91.4}/7.4}&\multirow{2}{*}{3.2/\textbf{85.1}/11.7}\\
        &  B  & \checkmark & \checkmark   &          &   \textbf{0.24}     & \textbf{83.3}         &      &           \\
      \hline
      \multirow{2}{*}{Abl.3}&A&\checkmark& * &&0.68 & \textbf{94.7} &\multirow{2}{*}{5.6/\textbf{84.3}/10.1}&\multirow{2}{*}{2.2/\textbf{86.6}/11.2} \\
        &  B  & \checkmark &\checkmark&  &  \textbf{0.24}    & 83.3           &     &           \\
      \hline
      \multirow{2}{*}{Abl.4}&A&\checkmark&\checkmark&&0.24 & 83.3  &\multirow{2}{*}{5.0/\textbf{76.0}/19.0}&\multirow{2}{*}{4.0/\textbf{69.0}/27.0}\\
        &  B  & \checkmark &\checkmark& \checkmark  &  \textbf{0.23}      & \textbf{85.5}         &     &           \\
      \hline
    \end{tabular}
    \vspace{-0.2cm}
    \caption{\textbf{Ablation study.} C, D, and U denote contact candidates, the human pose discriminator, and the contact candidate update, respectively. $P$ is penetration distance. $C_{acc}$ is contact accuracy.}
    \vspace{-0.3cm}
    \label{tab:results_ablation}
\end{table}

\textbf{User Studies.}
We conduct user studies to examine the quality of \dataset-Synthetic in terms of naturalness of contact and human motion.
Each study comprises two collections, each with at least 100 sequences displayed in pairs on a website. Users are instructed to assess the realism of human-object contacts and the naturalness of human motions, and then select the superior one in each pair separately. Recognizing the diversity of acceptable contacts and motions, participants are permitted to deem the performances as roughly equivalent.


 \noindent\textbf{Ablation on Contact Candidates.} In Table \ref{tab:results_ablation}.Abl.1, we only use the contact points from a source trajectory for retargeting to the target instead of resorting to the CORE4D-Real for many candidates,
 making the whole retargeting process similar to the OakInk~\cite{yang2022oakink} method. We observe a sharp decline in both physical plausibility and user preferences, indicating that our method compensates for OakInk's shortcomings in retargeting objects with significant geometric and scale variations.

\noindent\textbf{Ablations on Discriminator.} In Table \ref{tab:results_ablation}.Abl.2, we omit the human pose discriminator in the collaboration retargeting. Method A randomly chooses a candidate from the contact candidates. There are obvious performance drops, demonstrating the critical role of the human pose discriminator in selecting appropriate candidates. Table \ref{tab:results_ablation}.Abl.3 further compares the proposed discriminator against selecting the motion with the most accurate contact (method A), and user studies reveal significant superiority of the discriminator.

\noindent\textbf{Ablation on Contact Candidate Update.} We exclude contact candidate update process in Table \ref{tab:results_ablation}.Abl.4 experiment. This removal has weakened our method's ability to search for optimal solutions on objects, resulting in a modest degradation in penetration distance. The user study still exhibited a strong bias, indicating a perceived decline in the plausibility of both contact and motion. This ablation underscores the importance of contact candidate updates within our methodology.

\noindent\textbf{Comparing \dataset-Synthetic with \dataset-Real.} We assess the quality of CORE4D-Synthetic by comparing it with CORE4D-Real through user study. In conclusion, there is a 43\% probability that users perceive the quality of both options as comparable. Furthermore, in 14\% of cases, users even exhibit a preference for synthetic data. This indicates that the quality of our synthetic data closely approximates that of real data.



\section{Dataset Applications}
\subsection{\dataset-Synthetic Enhances Human-object Motion Forecasting Quality}

Table \ref{tab:results_downstream_application} compares the motion forecasting ability of light-weighted CAHMP~\cite{corona2020contextaware}. The test set is S2 defined in Section~\ref{sec:data_split}. We assess the quality of \dataset-Synthetic by comparing No.A and No.B. No.A even have better performance on object due to enriched spacial relation between human and object in \dataset-Synthetic. No.C shows the value of the \dataset-Synthetic by largely improving the performance. Details are in the appendix.


\begin{table}[tb]
  \centering
  \scriptsize
  \small
  \begin{tabular}{|c|c|c|c|c|c|c|}
    \hline
    \multirow{2}{*}{No} & \multicolumn{3}{c|}{Train Set} & Human & \multicolumn{2}{c|}{Object}  \\
    \cline{2-7}
    &Total &Real & Synthetic & $J_e$ ($\downarrow$) & $T_e$ ($\downarrow$) & $R_e$ ($\downarrow$) \\
    \hline
    A &1.0K & 0.1K & 0.9K & 127.7 & 121.7 & 8.04  \\
    \hline
    B &1.0K & 1.0K & 0    & 127.0 & 120.5 & 9.48  \\
    \hline
    C &5.0K & 1.0K & 4.0K & \textbf{116.2} & \textbf{112.1} & \textbf{6.99}  \\
  \hline
  \end{tabular}
  \vspace{-0.2cm}
  \caption{\textbf{Ablation on the incorporation of \dataset-Synthetic on the motion forecasting task.}}
  \vspace{-0.4cm}
  \label{tab:results_downstream_application}
\end{table}

\subsection{\dataset Supports Humanoid Skill Learning}


Benefitting from rapid developments of humanoid robots \cite{feng2014optimization,kojima2015development,unitree_h1}, tremendous progress has been made in studying versatile humanoid skills for locomotion \cite{he2024omnih2o,tang2024humanmimic,radosavovic2024humanoid,meng2023online} and humanoid-object interaction \cite{merel2020catch,dao2024sim,zhang2024wococo}. Aiming at enabling humanoid learning skills from human data, we select human interaction motions with three large boxes from \dataset, and retarget them onto the Unitree H1 humanoid robot \cite{unitree_h1} with object scale augmentation. With 890 human-like humanoid-box interaction sequences, we design a box-lifting task in Isaac Gym \cite{makoviychuk2021isaac}, and benchmark two state-of-the-art humanoid imitation learning (IL) methodologies \cite{ACT,fu2024humanplus} comparing to demonstration-free reinforcement learning (RL) paradigm \cite{rl_ppo}.

\begin{figure}[tb]
  \centering
  \includegraphics[width=1.0\linewidth]{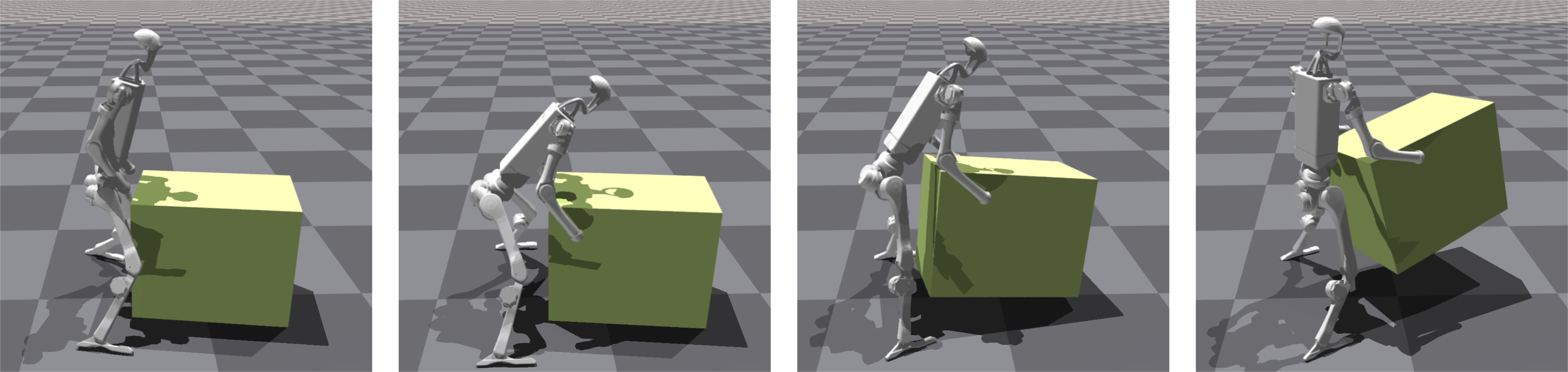}
  \vspace{-0.6cm}
  \caption{\textbf{Visualization of the humanoid box-lifting skill trained by \dataset via imitation learning.}}
  \vspace{-0.4cm}
  \label{fig:humanoid_box_lifting}
\end{figure}

Table \ref{tab:humanoid_box_lifting} compares the success rates of these methods. Leveraging interaction data from \dataset, the two IL methods~\cite{ACT,fu2024humanplus} consistently make it possible for humanoids to lift unseen boxes with visual sensor signals successfully, demonstrating that \dataset can promote humanoid interaction skill learning. Figure \ref{fig:humanoid_box_lifting} exemplifies a successful case of HumanPlus~\cite{fu2024humanplus}, showing that humanoids can learn from \dataset and achieve the task in a human-like manner. As the development of multi-humanoid imitation learning methods in the future, we anticipate that \dataset can further promote collaboration skill learning. Details on task formulation, method designs, and evaluations are in the appendix.

\begin{table}[tb]
  \centering
  \footnotesize
  \small
  \begin{tabular}{|c|c|c|c|}
  \hline
  Data & N/A & \multicolumn{2}{c|}{CORE4D} \\
  \hline
  Method & PPO \cite{rl_ppo} & HumanPlus \cite{fu2024humanplus} & HST \cite{fu2024humanplus} + ACT \cite{ACT}  \\
  \hline
  $SR$ ($\uparrow$) & 0.0 & 21.0 & 26.5 \\
  \hline
  \end{tabular}
  \vspace{-0.2cm}
  \caption{\textbf{Success rates of RL and IL in humanoid box lifting.}}
  \vspace{-0.3cm}
  \label{tab:humanoid_box_lifting}
\end{table}

\section{Conclusion and Limitations}
\label{sec:limitation}
We present \dataset, a novel large-scale 4D human-object-human interaction dataset for collaborative object rearrangement. It comprises diverse compositions of various object geometries, collaboration modes, and surrounding 3D scenes. To efficiently enlarge the data scale, we contribute a hybrid data acquisition method involving real-world data capturing and a novel synthetic data augmentation algorithm, resulting in 11K motion sequences covering 37 real-world and 3K virtual objects. Extensive experiments demonstrate the effectiveness of the data augmentation strategy and the value of the augmented motion data. We benchmark human-object motion forecasting and interaction synthesis on \dataset, revealing new challenges and research opportunities.

\noindent\textbf{Limitations.} Firstly, outdoor scenes are not incorporated in CORE4D-Real due to the usage of the mocap system. Secondly, visual signals are excluded in CORE4D-Synthetic. Transferring real-world videos onto synthesized collaboration motions could be an interesting future direction.



{
    \small
    \bibliographystyle{ieeenat_fullname}
    \bibliography{main}

\begin{thebibliography}{145}
\providecommand{\natexlab}[1]{#1}
\providecommand{\url}[1]{\texttt{#1}}
\expandafter\ifx\csname urlstyle\endcsname\relax
  \providecommand{\doi}[1]{doi: #1}\else
  \providecommand{\doi}{doi: \begingroup \urlstyle{rm}\Url}\fi

\bibitem[Adeli et~al.(2021)Adeli, Ehsanpour, Reid, Niebles, Savarese, Adeli, and Rezatofighi]{adeli2021tripod}
Vida Adeli, Mahsa Ehsanpour, Ian Reid, Juan~Carlos Niebles, Silvio Savarese, Ehsan Adeli, and Hamid Rezatofighi.
\newblock Tripod: Human trajectory and pose dynamics forecasting in the wild.
\newblock In \emph{Proceedings of the IEEE/CVF International Conference on Computer Vision}, pages 13390--13400, 2021.

\bibitem[Ara{\'u}jo et~al.(2023)Ara{\'u}jo, Li, Vetrivel, Agarwal, Wu, Gopinath, Clegg, and Liu]{CIRCLE}
Joao~Pedro Ara{\'u}jo, Jiaman Li, Karthik Vetrivel, Rishi Agarwal, Jiajun Wu, Deepak Gopinath, Alexander~William Clegg, and Karen Liu.
\newblock Circle: Capture in rich contextual environments.
\newblock In \emph{Proceedings of the IEEE/CVF Conference on Computer Vision and Pattern Recognition}, pages 21211--21221, 2023.

\bibitem[Bhatnagar et~al.(2022)Bhatnagar, Xie, Petrov, Sminchisescu, Theobalt, and Pons-Moll]{BEHAVE}
Bharat~Lal Bhatnagar, Xianghui Xie, Ilya~A Petrov, Cristian Sminchisescu, Christian Theobalt, and Gerard Pons-Moll.
\newblock Behave: Dataset and method for tracking human object interactions.
\newblock In \emph{Proceedings of the IEEE/CVF Conference on Computer Vision and Pattern Recognition}, pages 15935--15946, 2022.

\bibitem[Cao et~al.(2020)Cao, Gao, Mangalam, Cai, Vo, and Malik]{cao2020long}
Zhe Cao, Hang Gao, Karttikeya Mangalam, Qi-Zhi Cai, Minh Vo, and Jitendra Malik.
\newblock Long-term human motion prediction with scene context.
\newblock In \emph{Computer Vision--ECCV 2020: 16th European Conference, Glasgow, UK, August 23--28, 2020, Proceedings, Part I 16}, pages 387--404. Springer, 2020.

\bibitem[Carf{\`\i} et~al.(2019)Carf{\`\i}, Foglino, Bruno, and Mastrogiovanni]{carfi2019multi}
Alessandro Carf{\`\i}, Francesco Foglino, Barbara Bruno, and Fulvio Mastrogiovanni.
\newblock A multi-sensor dataset of human-human handover.
\newblock \emph{Data in brief}, 22:\penalty0 109--117, 2019.

\bibitem[Chan et~al.(2020)Chan, Pan, Croft, and Inaba]{chan2020affordance}
Wesley~P Chan, Matthew~KXJ Pan, Elizabeth~A Croft, and Masayuki Inaba.
\newblock An affordance and distance minimization based method for computing object orientations for robot human handovers.
\newblock \emph{International Journal of Social Robotics}, 12\penalty0 (1):\penalty0 143--162, 2020.

\bibitem[Chang et~al.(2015)Chang, Funkhouser, Guibas, Hanrahan, Huang, Li, Savarese, Savva, Song, Su, et~al.]{chang2015shapenet}
Angel~X Chang, Thomas Funkhouser, Leonidas Guibas, Pat Hanrahan, Qixing Huang, Zimo Li, Silvio Savarese, Manolis Savva, Shuran Song, Hao Su, et~al.
\newblock Shapenet: An information-rich 3d model repository.
\newblock \emph{arXiv preprint arXiv:1512.03012}, 2015.

\bibitem[Chao et~al.(2021)Chao, Yang, Xiang, Molchanov, Handa, Tremblay, Narang, Van~Wyk, Iqbal, Birchfield, et~al.]{chao2021dexycb}
Yu-Wei Chao, Wei Yang, Yu Xiang, Pavlo Molchanov, Ankur Handa, Jonathan Tremblay, Yashraj~S Narang, Karl Van~Wyk, Umar Iqbal, Stan Birchfield, et~al.
\newblock Dexycb: A benchmark for capturing hand grasping of objects.
\newblock In \emph{Proceedings of the IEEE/CVF Conference on Computer Vision and Pattern Recognition}, pages 9044--9053, 2021.

\bibitem[Chen et~al.(2022)Chen, Van~Wyk, Chao, Yang, Mousavian, Gupta, and Fox]{ISAGrasp}
Zoey~Qiuyu Chen, Karl Van~Wyk, Yu-Wei Chao, Wei Yang, Arsalan Mousavian, Abhishek Gupta, and Dieter Fox.
\newblock Learning robust real-world dexterous grasping policies via implicit shape augmentation.
\newblock \emph{arXiv preprint arXiv:2210.13638}, 2022.

\bibitem[Cheng et~al.(2023)Cheng, Oh, Price, Schwing, and Lee]{cheng2023tracking}
Ho~Kei Cheng, Seoung~Wug Oh, Brian Price, Alexander Schwing, and Joon-Young Lee.
\newblock Tracking anything with decoupled video segmentation.
\newblock In \emph{Proceedings of the IEEE/CVF International Conference on Computer Vision}, pages 1316--1326, 2023.

\bibitem[Christen et~al.(2022)Christen, Kocabas, Aksan, Hwangbo, Song, and Hilliges]{christen2022d}
Sammy Christen, Muhammed Kocabas, Emre Aksan, Jemin Hwangbo, Jie Song, and Otmar Hilliges.
\newblock D-grasp: Physically plausible dynamic grasp synthesis for hand-object interactions.
\newblock In \emph{Proceedings of the IEEE/CVF Conference on Computer Vision and Pattern Recognition}, pages 20577--20586, 2022.

\bibitem[Christen et~al.(2023)Christen, Yang, P{\'e}rez-D’Arpino, Hilliges, Fox, and Chao]{christen2023learning}
Sammy Christen, Wei Yang, Claudia P{\'e}rez-D’Arpino, Otmar Hilliges, Dieter Fox, and Yu-Wei Chao.
\newblock Learning human-to-robot handovers from point clouds.
\newblock In \emph{Proceedings of the IEEE/CVF Conference on Computer Vision and Pattern Recognition}, pages 9654--9664, 2023.

\bibitem[Cini et~al.(2019)Cini, Ortenzi, Corke, and Controzzi]{cini2019choice}
Francesca Cini, V Ortenzi, P Corke, and MJSR Controzzi.
\newblock On the choice of grasp type and location when handing over an object.
\newblock \emph{Science Robotics}, 4\penalty0 (27):\penalty0 eaau9757, 2019.

\bibitem[Corona et~al.(2020{\natexlab{a}})Corona, Pumarola, Alenya, and Moreno-Noguer]{CAHMP}
Enric Corona, Albert Pumarola, Guillem Alenya, and Francesc Moreno-Noguer.
\newblock Context-aware human motion prediction.
\newblock In \emph{Proceedings of the IEEE/CVF Conference on Computer Vision and Pattern Recognition}, pages 6992--7001, 2020{\natexlab{a}}.

\bibitem[Corona et~al.(2020{\natexlab{b}})Corona, Pumarola, Alenya, and Moreno-Noguer]{corona2020context}
Enric Corona, Albert Pumarola, Guillem Alenya, and Francesc Moreno-Noguer.
\newblock Context-aware human motion prediction.
\newblock In \emph{Proceedings of the IEEE/CVF Conference on Computer Vision and Pattern Recognition}, pages 6992--7001, 2020{\natexlab{b}}.

\bibitem[Corona et~al.(2020{\natexlab{c}})Corona, Pumarola, Alenyà, and Moreno-Noguer]{corona2020contextaware}
Enric Corona, Albert Pumarola, Guillem Alenyà, and Francesc Moreno-Noguer.
\newblock Context-aware human motion prediction, 2020{\natexlab{c}}.

\bibitem[Dabral et~al.(2021)Dabral, Shimada, Jain, Theobalt, and Golyanik]{GraviCap}
Rishabh Dabral, Soshi Shimada, Arjun Jain, Christian Theobalt, and Vladislav Golyanik.
\newblock Gravity-aware monocular 3d human-object reconstruction.
\newblock In \emph{Proceedings of the IEEE/CVF International Conference on Computer Vision}, pages 12365--12374, 2021.

\bibitem[Dai et~al.(2022)Dai, Lin, Wen, Shen, Xu, Yu, Ma, and Wang]{HSC4D}
Yudi Dai, Yitai Lin, Chenglu Wen, Siqi Shen, Lan Xu, Jingyi Yu, Yuexin Ma, and Cheng Wang.
\newblock Hsc4d: Human-centered 4d scene capture in large-scale indoor-outdoor space using wearable imus and lidar.
\newblock In \emph{Proceedings of the IEEE/CVF Conference on Computer Vision and Pattern Recognition}, pages 6792--6802, 2022.

\bibitem[Dai et~al.(2023)Dai, Lin, Lin, Wen, Xu, Yi, Shen, Ma, and Wang]{Sloper4D}
Yudi Dai, YiTai Lin, XiPing Lin, Chenglu Wen, Lan Xu, Hongwei Yi, Siqi Shen, Yuexin Ma, and Cheng Wang.
\newblock Sloper4d: A scene-aware dataset for global 4d human pose estimation in urban environments.
\newblock In \emph{Proceedings of the IEEE/CVF Conference on Computer Vision and Pattern Recognition}, pages 682--692, 2023.

\bibitem[Dao et~al.(2023)Dao, Duan, and Fern]{dao2023sim}
Jeremy Dao, Helei Duan, and Alan Fern.
\newblock Sim-to-real learning for humanoid box loco-manipulation.
\newblock \emph{arXiv preprint arXiv:2310.03191}, 2023.

\bibitem[Dao et~al.(2024)Dao, Duan, and Fern]{dao2024sim}
Jeremy Dao, Helei Duan, and Alan Fern.
\newblock Sim-to-real learning for humanoid box loco-manipulation.
\newblock In \emph{2024 IEEE International Conference on Robotics and Automation (ICRA)}, pages 16930--16936. IEEE, 2024.

\bibitem[Diller and Dai(2023)]{CG-HOI}
Christian Diller and Angela Dai.
\newblock Cg-hoi: Contact-guided 3d human-object interaction generation.
\newblock \emph{arXiv preprint arXiv:2311.16097}, 2023.

\bibitem[Fan et~al.(2024)Fan, Du, Cai, Peng, and Sun]{fan2024textim}
Siyuan Fan, Bo Du, Xiantao Cai, Bo Peng, and Longling Sun.
\newblock Textim: Part-aware interactive motion synthesis from text.
\newblock \emph{arXiv preprint arXiv:2408.03302}, 2024.

\bibitem[Fan et~al.(2023)Fan, Taheri, Tzionas, Kocabas, Kaufmann, Black, and Hilliges]{ARCTIC}
Zicong Fan, Omid Taheri, Dimitrios Tzionas, Muhammed Kocabas, Manuel Kaufmann, Michael~J Black, and Otmar Hilliges.
\newblock Arctic: A dataset for dexterous bimanual hand-object manipulation.
\newblock In \emph{Proceedings of the IEEE/CVF Conference on Computer Vision and Pattern Recognition}, pages 12943--12954, 2023.

\bibitem[Feng et~al.(2014)Feng, Whitman, Xinjilefu, and Atkeson]{feng2014optimization}
Siyuan Feng, Eric Whitman, X Xinjilefu, and Christopher~G Atkeson.
\newblock Optimization based full body control for the atlas robot.
\newblock In \emph{2014 IEEE-RAS International Conference on Humanoid Robots}, pages 120--127. IEEE, 2014.

\bibitem[Fu et~al.(2024)Fu, Zhao, Wu, Wetzstein, and Finn]{fu2024humanplus}
Zipeng Fu, Qingqing Zhao, Qi Wu, Gordon Wetzstein, and Chelsea Finn.
\newblock Humanplus: Humanoid shadowing and imitation from humans.
\newblock \emph{arXiv preprint arXiv:2406.10454}, 2024.

\bibitem[Ghosh et~al.(2023)Ghosh, Dabral, Golyanik, Theobalt, and Slusallek]{IMOS}
Anindita Ghosh, Rishabh Dabral, Vladislav Golyanik, Christian Theobalt, and Philipp Slusallek.
\newblock Imos: Intent-driven full-body motion synthesis for human-object interactions.
\newblock In \emph{Computer Graphics Forum}, pages 1--12. Wiley Online Library, 2023.

\bibitem[Guo et~al.(2022)Guo, Bie, Alameda-Pineda, and Moreno-Noguer]{guo2022multi}
Wen Guo, Xiaoyu Bie, Xavier Alameda-Pineda, and Francesc Moreno-Noguer.
\newblock Multi-person extreme motion prediction.
\newblock In \emph{Proceedings of the IEEE/CVF Conference on Computer Vision and Pattern Recognition}, pages 13053--13064, 2022.

\bibitem[Guo et~al.(2023)Guo, Du, Shen, Lepetit, Alameda-Pineda, and Moreno-Noguer]{guo2023back}
Wen Guo, Yuming Du, Xi Shen, Vincent Lepetit, Xavier Alameda-Pineda, and Francesc Moreno-Noguer.
\newblock Back to mlp: A simple baseline for human motion prediction.
\newblock In \emph{Proceedings of the IEEE/CVF Winter Conference on Applications of Computer Vision}, pages 4809--4819, 2023.

\bibitem[Guzov et~al.(2021)Guzov, Mir, Sattler, and Pons-Moll]{HPS}
Vladimir Guzov, Aymen Mir, Torsten Sattler, and Gerard Pons-Moll.
\newblock Human poseitioning system (hps): 3d human pose estimation and self-localization in large scenes from body-mounted sensors.
\newblock In \emph{Proceedings of the IEEE/CVF Conference on Computer Vision and Pattern Recognition}, pages 4318--4329, 2021.

\bibitem[Guzov et~al.(2022)Guzov, Chibane, Marin, He, Sattler, and Pons-Moll]{guzov2022interaction}
Vladimir Guzov, Julian Chibane, Riccardo Marin, Yannan He, Torsten Sattler, and Gerard Pons-Moll.
\newblock Interaction replica: Tracking human-object interaction and scene changes from human motion.
\newblock \emph{arXiv preprint arXiv:2205.02830}, 2022.

\bibitem[Hassan et~al.(2019)Hassan, Choutas, Tzionas, and Black]{PROX}
Mohamed Hassan, Vasileios Choutas, Dimitrios Tzionas, and Michael~J Black.
\newblock Resolving 3d human pose ambiguities with 3d scene constraints.
\newblock In \emph{Proceedings of the IEEE/CVF international conference on computer vision}, pages 2282--2292, 2019.

\bibitem[Hassan et~al.(2021{\natexlab{a}})Hassan, Ceylan, Villegas, Saito, Yang, Zhou, and Black]{SAMP}
Mohamed Hassan, Duygu Ceylan, Ruben Villegas, Jun Saito, Jimei Yang, Yi Zhou, and Michael~J Black.
\newblock Stochastic scene-aware motion prediction.
\newblock In \emph{Proceedings of the IEEE/CVF International Conference on Computer Vision}, pages 11374--11384, 2021{\natexlab{a}}.

\bibitem[Hassan et~al.(2021{\natexlab{b}})Hassan, Ghosh, Tesch, Tzionas, and Black]{POSA}
Mohamed Hassan, Partha Ghosh, Joachim Tesch, Dimitrios Tzionas, and Michael~J Black.
\newblock Populating 3d scenes by learning human-scene interaction.
\newblock In \emph{Proceedings of the IEEE/CVF Conference on Computer Vision and Pattern Recognition}, pages 14708--14718, 2021{\natexlab{b}}.

\bibitem[He et~al.(2024{\natexlab{a}})He, Luo, He, Xiao, Zhang, Zhang, Kitani, Liu, and Shi]{he2024omnih2o}
Tairan He, Zhengyi Luo, Xialin He, Wenli Xiao, Chong Zhang, Weinan Zhang, Kris Kitani, Changliu Liu, and Guanya Shi.
\newblock Omnih2o: Universal and dexterous human-to-humanoid whole-body teleoperation and learning.
\newblock \emph{arXiv preprint arXiv:2406.08858}, 2024{\natexlab{a}}.

\bibitem[He et~al.(2024{\natexlab{b}})He, Luo, Xiao, Zhang, Kitani, Liu, and Shi]{he2024learning}
Tairan He, Zhengyi Luo, Wenli Xiao, Chong Zhang, Kris Kitani, Changliu Liu, and Guanya Shi.
\newblock Learning human-to-humanoid real-time whole-body teleoperation.
\newblock \emph{arXiv preprint arXiv:2403.04436}, 2024{\natexlab{b}}.

\bibitem[Ho et~al.(2020)Ho, Jain, and Abbeel]{ho2020denoising}
Jonathan Ho, Ajay Jain, and Pieter Abbeel.
\newblock Denoising diffusion probabilistic models.
\newblock \emph{Advances in neural information processing systems}, 33:\penalty0 6840--6851, 2020.

\bibitem[Huang et~al.(2022{\natexlab{a}})Huang, Yi, H{\"o}schle, Safroshkin, Alexiadis, Polikovsky, Scharstein, and Black]{RICH}
Chun-Hao~P Huang, Hongwei Yi, Markus H{\"o}schle, Matvey Safroshkin, Tsvetelina Alexiadis, Senya Polikovsky, Daniel Scharstein, and Michael~J Black.
\newblock Capturing and inferring dense full-body human-scene contact.
\newblock In \emph{Proceedings of the IEEE/CVF Conference on Computer Vision and Pattern Recognition}, pages 13274--13285, 2022{\natexlab{a}}.

\bibitem[Huang et~al.(2022{\natexlab{b}})Huang, Taheri, Black, and Tzionas]{InterCap}
Yinghao Huang, Omid Taheri, Michael~J Black, and Dimitrios Tzionas.
\newblock Intercap: Joint markerless 3d tracking of humans and objects in interaction.
\newblock In \emph{DAGM German Conference on Pattern Recognition}, pages 281--299. Springer, 2022{\natexlab{b}}.

\bibitem[Huang et~al.(2024)Huang, Xu, Huang, Ma, Huang, and Hu]{huang2024spatial}
Zeyu Huang, Honghao Xu, Haibin Huang, Chongyang Ma, Hui Huang, and Ruizhen Hu.
\newblock Spatial and surface correspondence field for interaction transfer.
\newblock \emph{arXiv preprint arXiv:2405.03221}, 2024.

\bibitem[Jiang et~al.(2023)Jiang, Liu, Cao, Cui, Zhang, Chen, Wang, Zhu, and Huang]{CHAIRS}
Nan Jiang, Tengyu Liu, Zhexuan Cao, Jieming Cui, Zhiyuan Zhang, Yixin Chen, He Wang, Yixin Zhu, and Siyuan Huang.
\newblock Full-body articulated human-object interaction.
\newblock In \emph{Proceedings of the IEEE/CVF International Conference on Computer Vision}, pages 9365--9376, 2023.

\bibitem[Jiang et~al.(2024{\natexlab{a}})Jiang, He, Wang, Li, Chen, Huang, and Zhu]{jiang2024autonomous}
Nan Jiang, Zimo He, Zi Wang, Hongjie Li, Yixin Chen, Siyuan Huang, and Yixin Zhu.
\newblock Autonomous character-scene interaction synthesis from text instruction.
\newblock \emph{arXiv preprint arXiv:2410.03187}, 2024{\natexlab{a}}.

\bibitem[Jiang et~al.(2024{\natexlab{b}})Jiang, Zhang, Li, Ma, Wang, Chen, Liu, Zhu, and Huang]{jiang2024scaling}
Nan Jiang, Zhiyuan Zhang, Hongjie Li, Xiaoxuan Ma, Zan Wang, Yixin Chen, Tengyu Liu, Yixin Zhu, and Siyuan Huang.
\newblock Scaling up dynamic human-scene interaction modeling.
\newblock \emph{arXiv preprint arXiv:2403.08629}, 2024{\natexlab{b}}.

\bibitem[Kasahara et~al.(2017)Kasahara, Konno, Owaki, Nishi, Takeshita, Ito, Kasuga, and Ushiba]{kasahara2017malleable}
Shunichi Kasahara, Keina Konno, Richi Owaki, Tsubasa Nishi, Akiko Takeshita, Takayuki Ito, Shoko Kasuga, and Junichi Ushiba.
\newblock Malleable embodiment: changing sense of embodiment by spatial-temporal deformation of virtual human body.
\newblock In \emph{Proceedings of the 2017 CHI Conference on Human Factors in Computing Systems}, pages 6438--6448, 2017.

\bibitem[Khanna et~al.(2023)Khanna, Bj{\"o}rkman, and Smith]{khanna2023multimodal}
Parag Khanna, M{\aa}rten Bj{\"o}rkman, and Christian Smith.
\newblock A multimodal data set of human handovers with design implications for human-robot handovers.
\newblock In \emph{2023 32nd IEEE International Conference on Robot and Human Interactive Communication (RO-MAN)}, pages 1843--1850. IEEE, 2023.

\bibitem[Kim et~al.(2016)Kim, Park, Bang, and Lee]{kim2016retargeting}
Yeonjoon Kim, Hangil Park, Seungbae Bang, and Sung-Hee Lee.
\newblock Retargeting human-object interaction to virtual avatars.
\newblock \emph{IEEE transactions on visualization and computer graphics}, 22\penalty0 (11):\penalty0 2405--2412, 2016.

\bibitem[Kojima et~al.(2015)Kojima, Karasawa, Kozuki, Kuroiwa, Yukizaki, Iwaishi, Ishikawa, Koyama, Noda, Sugai, et~al.]{kojima2015development}
Kunio Kojima, Tatsuhi Karasawa, Toyotaka Kozuki, Eisoku Kuroiwa, Sou Yukizaki, Satoshi Iwaishi, Tatsuya Ishikawa, Ryo Koyama, Shintaro Noda, Fumihito Sugai, et~al.
\newblock Development of life-sized high-power humanoid robot jaxon for real-world use.
\newblock In \emph{2015 IEEE-RAS 15th International Conference on Humanoid Robots (Humanoids)}, pages 838--843. IEEE, 2015.

\bibitem[Kshirsagar et~al.(2023)Kshirsagar, Fortuna, Xie, and Hoffman]{kshirsagar2023dataset}
Alap Kshirsagar, Raphael Fortuna, Zhiming Xie, and Guy Hoffman.
\newblock Dataset of bimanual human-to-human object handovers.
\newblock \emph{Data in Brief}, 48:\penalty0 109277, 2023.

\bibitem[Kulkarni et~al.(2023)Kulkarni, Rempe, Genova, Kundu, Johnson, Fouhey, and Guibas]{NIFTY}
Nilesh Kulkarni, Davis Rempe, Kyle Genova, Abhijit Kundu, Justin Johnson, David Fouhey, and Leonidas Guibas.
\newblock Nifty: Neural object interaction fields for guided human motion synthesis.
\newblock \emph{arXiv preprint arXiv:2307.07511}, 2023.

\bibitem[Li and Nguyen(2023)]{li2023kinodynamics}
Junheng Li and Quan Nguyen.
\newblock Kinodynamics-based pose optimization for humanoid loco-manipulation.
\newblock \emph{arXiv preprint arXiv:2303.04985}, 2023.

\bibitem[Li et~al.(2023{\natexlab{a}})Li, Bian, Xu, Chen, Yang, and Lu]{li2023hybrik}
Jiefeng Li, Siyuan Bian, Chao Xu, Zhicun Chen, Lixin Yang, and Cewu Lu.
\newblock Hybrik-x: Hybrid analytical-neural inverse kinematics for whole-body mesh recovery.
\newblock \emph{arXiv preprint arXiv:2304.05690}, 2023{\natexlab{a}}.

\bibitem[Li et~al.(2023{\natexlab{b}})Li, Clegg, Mottaghi, Wu, Puig, and Liu]{CHOIS}
Jiaman Li, Alexander Clegg, Roozbeh Mottaghi, Jiajun Wu, Xavier Puig, and C~Karen Liu.
\newblock Controllable human-object interaction synthesis.
\newblock \emph{arXiv preprint arXiv:2312.03913}, 2023{\natexlab{b}}.

\bibitem[Li et~al.(2023{\natexlab{c}})Li, Wu, and Liu]{OMOMO}
Jiaman Li, Jiajun Wu, and C~Karen Liu.
\newblock Object motion guided human motion synthesis.
\newblock \emph{ACM Transactions on Graphics (TOG)}, 42\penalty0 (6):\penalty0 1--11, 2023{\natexlab{c}}.

\bibitem[Li et~al.(2022)Li, Shimada, Schiele, Theobalt, and Golyanik]{MocapDeform}
Zhi Li, Soshi Shimada, Bernt Schiele, Christian Theobalt, and Vladislav Golyanik.
\newblock Mocapdeform: Monocular 3d human motion capture in deformable scenes.
\newblock In \emph{2022 International Conference on 3D Vision (3DV)}, pages 1--11. IEEE, 2022.

\bibitem[Liu et~al.(2021)Liu, Yang, Zhang, Cui, Rehg, and Tang]{4DCapture}
Miao Liu, Dexin Yang, Yan Zhang, Zhaopeng Cui, James~M Rehg, and Siyu Tang.
\newblock 4d human body capture from egocentric video via 3d scene grounding.
\newblock In \emph{2021 international conference on 3D vision (3DV)}, pages 930--939. IEEE, 2021.

\bibitem[Liu et~al.(2024{\natexlab{a}})Liu, Li, Fang, Liu, You, and Lu]{P3HAOI}
Siqi Liu, Yong-Lu Li, Zhou Fang, Xinpeng Liu, Yang You, and Cewu Lu.
\newblock Primitive-based 3d human-object interaction modelling and programming.
\newblock In \emph{Proceedings of the AAAI Conference on Artificial Intelligence}, pages 3711--3719, 2024{\natexlab{a}}.

\bibitem[Liu et~al.(2022)Liu, Liu, Jiang, Lyu, Wan, Shen, Liang, Fu, Wang, and Yi]{HOI4D}
Yunze Liu, Yun Liu, Che Jiang, Kangbo Lyu, Weikang Wan, Hao Shen, Boqiang Liang, Zhoujie Fu, He Wang, and Li Yi.
\newblock Hoi4d: A 4d egocentric dataset for category-level human-object interaction.
\newblock In \emph{Proceedings of the IEEE/CVF Conference on Computer Vision and Pattern Recognition}, pages 21013--21022, 2022.

\bibitem[Liu et~al.(2023)Liu, Chen, and Yi]{InteractiveHumanoid}
Yunze Liu, Changxi Chen, and Li Yi.
\newblock Interactive humanoid: Online full-body motion reaction synthesis with social affordance canonicalization and forecasting.
\newblock \emph{arXiv preprint arXiv:2312.08983}, 2023.

\bibitem[Liu et~al.(2024{\natexlab{b}})Liu, Yang, Si, Liu, Li, Zhang, Liu, and Yi]{TACO}
Yun Liu, Haolin Yang, Xu Si, Ling Liu, Zipeng Li, Yuxiang Zhang, Yebin Liu, and Li Yi.
\newblock Taco: Benchmarking generalizable bimanual tool-action-object understanding.
\newblock In \emph{Proceedings of the IEEE/CVF Conference on Computer Vision and Pattern Recognition}, pages 21740--21751, 2024{\natexlab{b}}.

\bibitem[Liu et~al.(2024{\natexlab{c}})Liu, Yang, Si, Liu, Li, Zhang, Liu, and Yi]{liu2024taco}
Yun Liu, Haolin Yang, Xu Si, Ling Liu, Zipeng Li, Yuxiang Zhang, Yebin Liu, and Li Yi.
\newblock Taco: Benchmarking generalizable bimanual tool-action-object understanding.
\newblock \emph{arXiv preprint arXiv:2401.08399}, 2024{\natexlab{c}}.

\bibitem[Lorensen and Cline(1998)]{lorensen1998marching}
William~E Lorensen and Harvey~E Cline.
\newblock Marching cubes: A high resolution 3d surface construction algorithm.
\newblock In \emph{Seminal graphics: pioneering efforts that shaped the field}, pages 347--353. 1998.

\bibitem[Luo et~al.(2023)Luo, Cao, Kitani, Xu, et~al.]{phc}
Zhengyi Luo, Jinkun Cao, Kris Kitani, Weipeng Xu, et~al.
\newblock Perpetual humanoid control for real-time simulated avatars.
\newblock In \emph{Proceedings of the IEEE/CVF International Conference on Computer Vision}, pages 10895--10904, 2023.

\bibitem[Lv et~al.(2025)Lv, Xu, Yan, Jin, Xu, Wu, Liu, Li, Bi, Zeng, et~al.]{lv2025himo}
Xintao Lv, Liang Xu, Yichao Yan, Xin Jin, Congsheng Xu, Shuwen Wu, Yifan Liu, Lincheng Li, Mengxiao Bi, Wenjun Zeng, et~al.
\newblock Himo: A new benchmark for full-body human interacting with multiple objects.
\newblock In \emph{European Conference on Computer Vision}, pages 300--318. Springer, 2025.

\bibitem[Makoviychuk et~al.(2021)Makoviychuk, Wawrzyniak, Guo, Lu, Storey, Macklin, Hoeller, Rudin, Allshire, Handa, et~al.]{makoviychuk2021isaac}
Viktor Makoviychuk, Lukasz Wawrzyniak, Yunrong Guo, Michelle Lu, Kier Storey, Miles Macklin, David Hoeller, Nikita Rudin, Arthur Allshire, Ankur Handa, et~al.
\newblock Isaac gym: High performance gpu-based physics simulation for robot learning.
\newblock \emph{arXiv preprint arXiv:2108.10470}, 2021.

\bibitem[Mao et~al.(2022)Mao, Hartley, Salzmann, et~al.]{mao2022contact}
Wei Mao, Richard~I Hartley, Mathieu Salzmann, et~al.
\newblock Contact-aware human motion forecasting.
\newblock \emph{Advances in Neural Information Processing Systems}, 35:\penalty0 7356--7367, 2022.

\bibitem[Meng et~al.(2023)Meng, Yu, Chen, Huang, Meng, and Huang]{meng2023online}
Xiang Meng, Zhangguo Yu, Xuechao Chen, Zelin Huang, Fei Meng, and Qiang Huang.
\newblock Online adaptive motion generation for humanoid locomotion on non-flat terrain via template behavior extension.
\newblock \emph{IEEE Transactions on Automation Science and Engineering}, 2023.

\bibitem[Meredith et~al.(2001)Meredith, Maddock, et~al.]{meredith2001motion}
Maddock Meredith, Steve Maddock, et~al.
\newblock Motion capture file formats explained.
\newblock \emph{Department of Computer Science, University of Sheffield}, 211:\penalty0 241--244, 2001.

\bibitem[Merel et~al.(2020)Merel, Tunyasuvunakool, Ahuja, Tassa, Hasenclever, Pham, Erez, Wayne, and Heess]{merel2020catch}
Josh Merel, Saran Tunyasuvunakool, Arun Ahuja, Yuval Tassa, Leonard Hasenclever, Vu Pham, Tom Erez, Greg Wayne, and Nicolas Heess.
\newblock Catch \& carry: reusable neural controllers for vision-guided whole-body tasks.
\newblock \emph{ACM Transactions on Graphics (TOG)}, 39\penalty0 (4):\penalty0 39--1, 2020.

\bibitem[Mildenhall et~al.(2021)Mildenhall, Srinivasan, Tancik, Barron, Ramamoorthi, and Ng]{mildenhall2021nerf}
Ben Mildenhall, Pratul~P Srinivasan, Matthew Tancik, Jonathan~T Barron, Ravi Ramamoorthi, and Ren Ng.
\newblock Nerf: Representing scenes as neural radiance fields for view synthesis.
\newblock \emph{Communications of the ACM}, 65\penalty0 (1):\penalty0 99--106, 2021.

\bibitem[Mir et~al.(2024)Mir, Puig, Kanazawa, and Pons-Moll]{mir2024generating}
Aymen Mir, Xavier Puig, Angjoo Kanazawa, and Gerard Pons-Moll.
\newblock Generating continual human motion in diverse 3d scenes.
\newblock In \emph{2024 International Conference on 3D Vision (3DV)}, pages 903--913. IEEE, 2024.

\bibitem[Murooka et~al.(2021)Murooka, Kumagai, Morisawa, Kanehiro, and Kheddar]{murooka2021humanoid}
Masaki Murooka, Iori Kumagai, Mitsuharu Morisawa, Fumio Kanehiro, and Abderrahmane Kheddar.
\newblock Humanoid loco-manipulation planning based on graph search and reachability maps.
\newblock \emph{IEEE Robotics and Automation Letters}, 6\penalty0 (2):\penalty0 1840--1847, 2021.

\bibitem[Ng et~al.(2023{\natexlab{a}})Ng, Liu, and Kennedy]{ng2023diffusion}
Eley Ng, Ziang Liu, and Monroe Kennedy.
\newblock Diffusion co-policy for synergistic human-robot collaborative tasks.
\newblock \emph{IEEE Robotics and Automation Letters}, 2023{\natexlab{a}}.

\bibitem[Ng et~al.(2023{\natexlab{b}})Ng, Liu, and Kennedy]{ng2023takes}
Eley Ng, Ziang Liu, and Monroe Kennedy.
\newblock It takes two: Learning to plan for human-robot cooperative carrying.
\newblock In \emph{2023 IEEE International Conference on Robotics and Automation (ICRA)}, pages 7526--7532. IEEE, 2023{\natexlab{b}}.

\bibitem[NOITOM~INTERNATIONAL(2024)]{noitom}
INC NOITOM~INTERNATIONAL.
\newblock Noitom motion capture systems.
\newblock \url{https://www.noitom.com.cn/}, 2024.

\bibitem[OpenAI(2023)]{ChatGPT}
OpenAI.
\newblock Chatgpt.
\newblock \url{https://chat.openai.com/}, 2023.

\bibitem[OpenCV(2013)]{haarcascades}
OpenCV.
\newblock opencv.
\newblock \url{https://github.com/opencv/opencv/blob/master/data/haarcascades/haarcascade_frontalface_default.xml}, 2013.

\bibitem[Ouyang et~al.(2022)Ouyang, Wu, Jiang, Almeida, Wainwright, Mishkin, Zhang, Agarwal, Slama, Ray, et~al.]{RLHF}
Long Ouyang, Jeffrey Wu, Xu Jiang, Diogo Almeida, Carroll Wainwright, Pamela Mishkin, Chong Zhang, Sandhini Agarwal, Katarina Slama, Alex Ray, et~al.
\newblock Training language models to follow instructions with human feedback.
\newblock \emph{Advances in Neural Information Processing Systems}, 35:\penalty0 27730--27744, 2022.

\bibitem[Park et~al.(2019)Park, Florence, Straub, Newcombe, and Lovegrove]{park2019deepsdf}
Jeong~Joon Park, Peter Florence, Julian Straub, Richard Newcombe, and Steven Lovegrove.
\newblock Deepsdf: Learning continuous signed distance functions for shape representation.
\newblock In \emph{Proceedings of the IEEE/CVF conference on computer vision and pattern recognition}, pages 165--174, 2019.

\bibitem[Pavlakos et~al.(2019)Pavlakos, Choutas, Ghorbani, Bolkart, Osman, Tzionas, and Black]{SMPLX}
Georgios Pavlakos, Vasileios Choutas, Nima Ghorbani, Timo Bolkart, Ahmed~AA Osman, Dimitrios Tzionas, and Michael~J Black.
\newblock Expressive body capture: 3d hands, face, and body from a single image.
\newblock In \emph{Proceedings of the IEEE/CVF conference on computer vision and pattern recognition}, pages 10975--10985, 2019.

\bibitem[Peng et~al.(2022)Peng, Shen, Wang, Nie, Wang, and Wu]{peng2022somoformer}
Xiaogang Peng, Yaodi Shen, Haoran Wang, Binling Nie, Yigang Wang, and Zizhao Wu.
\newblock Somoformer: Social-aware motion transformer for multi-person motion prediction.
\newblock \emph{arXiv preprint arXiv:2208.09224}, 2022.

\bibitem[Peng et~al.(2023)Peng, Xie, Wu, Jampani, Sun, and Jiang]{HOI-Diff}
Xiaogang Peng, Yiming Xie, Zizhao Wu, Varun Jampani, Deqing Sun, and Huaizu Jiang.
\newblock Hoi-diff: Text-driven synthesis of 3d human-object interactions using diffusion models.
\newblock \emph{arXiv preprint arXiv:2312.06553}, 2023.

\bibitem[Peng et~al.(2018)Peng, Abbeel, Levine, and Van~de Panne]{peng2018deepmimic}
Xue~Bin Peng, Pieter Abbeel, Sergey Levine, and Michiel Van~de Panne.
\newblock Deepmimic: Example-guided deep reinforcement learning of physics-based character skills.
\newblock \emph{ACM Transactions On Graphics (TOG)}, 37\penalty0 (4):\penalty0 1--14, 2018.

\bibitem[Radford et~al.(2021)Radford, Kim, Hallacy, Ramesh, Goh, Agarwal, Sastry, Askell, Mishkin, Clark, et~al.]{radford2021learning}
Alec Radford, Jong~Wook Kim, Chris Hallacy, Aditya Ramesh, Gabriel Goh, Sandhini Agarwal, Girish Sastry, Amanda Askell, Pamela Mishkin, Jack Clark, et~al.
\newblock Learning transferable visual models from natural language supervision.
\newblock In \emph{International conference on machine learning}, pages 8748--8763. PMLR, 2021.

\bibitem[Radosavovic et~al.(2024)Radosavovic, Zhang, Shi, Rajasegaran, Kamat, Darrell, Sreenath, and Malik]{radosavovic2024humanoid}
Ilija Radosavovic, Bike Zhang, Baifeng Shi, Jathushan Rajasegaran, Sarthak Kamat, Trevor Darrell, Koushil Sreenath, and Jitendra Malik.
\newblock Humanoid locomotion as next token prediction.
\newblock \emph{arXiv preprint arXiv:2402.19469}, 2024.

\bibitem[Rodriguez and Behnke(2018)]{rodriguez2018transferring}
Diego Rodriguez and Sven Behnke.
\newblock Transferring category-based functional grasping skills by latent space non-rigid registration.
\newblock \emph{IEEE Robotics and Automation Letters}, 3\penalty0 (3):\penalty0 2662--2669, 2018.

\bibitem[Savva et~al.(2016)Savva, Chang, Hanrahan, Fisher, and Nie{\ss}ner]{savva2016pigraphs}
Manolis Savva, Angel~X Chang, Pat Hanrahan, Matthew Fisher, and Matthias Nie{\ss}ner.
\newblock Pigraphs: learning interaction snapshots from observations.
\newblock \emph{ACM Transactions On Graphics (TOG)}, 35\penalty0 (4):\penalty0 1--12, 2016.

\bibitem[Schulman et~al.(2017)Schulman, Wolski, Dhariwal, Radford, and Klimov]{rl_ppo}
John Schulman, Filip Wolski, Prafulla Dhariwal, Alec Radford, and Oleg Klimov.
\newblock Proximal policy optimization algorithms.
\newblock \emph{arXiv preprint arXiv:1707.06347}, 2017.

\bibitem[Simeonov et~al.(2022)Simeonov, Du, Tagliasacchi, Tenenbaum, Rodriguez, Agrawal, and Sitzmann]{simeonov2022neural}
Anthony Simeonov, Yilun Du, Andrea Tagliasacchi, Joshua~B Tenenbaum, Alberto Rodriguez, Pulkit Agrawal, and Vincent Sitzmann.
\newblock Neural descriptor fields: Se (3)-equivariant object representations for manipulation.
\newblock In \emph{2022 International Conference on Robotics and Automation (ICRA)}, pages 6394--6400. IEEE, 2022.

\bibitem[Sohn et~al.(2015)Sohn, Lee, and Yan]{CVAE}
Kihyuk Sohn, Honglak Lee, and Xinchen Yan.
\newblock Learning structured output representation using deep conditional generative models.
\newblock \emph{Advances in neural information processing systems}, 28, 2015.

\bibitem[Song et~al.(2024)Song, Zhang, Li, Gao, Hao, Hou, Chen, Li, and Qin]{song2024hoianimator}
Wenfeng Song, Xinyu Zhang, Shuai Li, Yang Gao, Aimin Hao, Xia Hou, Chenglizhao Chen, Ning Li, and Hong Qin.
\newblock Hoianimator: Generating text-prompt human-object animations using novel perceptive diffusion models.
\newblock In \emph{Proceedings of the IEEE/CVF Conference on Computer Vision and Pattern Recognition}, pages 811--820, 2024.

\bibitem[Starke et~al.(2019)Starke, Zhang, Komura, and Saito]{NSM}
Sebastian Starke, He Zhang, Taku Komura, and Jun Saito.
\newblock Neural state machine for character-scene interactions.
\newblock \emph{ACM Trans. Graph.}, 38\penalty0 (6):\penalty0 209--1, 2019.

\bibitem[Taheri et~al.(2020)Taheri, Ghorbani, Black, and Tzionas]{GRAB}
Omid Taheri, Nima Ghorbani, Michael~J Black, and Dimitrios Tzionas.
\newblock Grab: A dataset of whole-body human grasping of objects.
\newblock In \emph{Computer Vision--ECCV 2020: 16th European Conference, Glasgow, UK, August 23--28, 2020, Proceedings, Part IV 16}, pages 581--600. Springer, 2020.

\bibitem[Taheri et~al.(2022)Taheri, Choutas, Black, and Tzionas]{GOAL}
Omid Taheri, Vasileios Choutas, Michael~J Black, and Dimitrios Tzionas.
\newblock Goal: Generating 4d whole-body motion for hand-object grasping.
\newblock In \emph{Proceedings of the IEEE/CVF Conference on Computer Vision and Pattern Recognition}, pages 13263--13273, 2022.

\bibitem[Tang et~al.(2024)Tang, Hiraoka, Hiraoka, Shi, Kawaharazuka, Kojima, Okada, and Inaba]{tang2024humanmimic}
Annan Tang, Takuma Hiraoka, Naoki Hiraoka, Fan Shi, Kento Kawaharazuka, Kunio Kojima, Kei Okada, and Masayuki Inaba.
\newblock Humanmimic: Learning natural locomotion and transitions for humanoid robot via wasserstein adversarial imitation.
\newblock In \emph{2024 IEEE International Conference on Robotics and Automation (ICRA)}, pages 13107--13114. IEEE, 2024.

\bibitem[Tanke et~al.(2024)Tanke, Kwon, Mueller, Doering, and Gall]{tanke2024humans}
Julian Tanke, Oh-Hun Kwon, Felix~B Mueller, Andreas Doering, and Juergen Gall.
\newblock Humans in kitchens: A dataset for multi-person human motion forecasting with scene context.
\newblock \emph{Advances in Neural Information Processing Systems}, 36, 2024.

\bibitem[Tessler et~al.(2024)Tessler, Guo, Nabati, Chechik, and Peng]{tessler2024maskedmimic}
Chen Tessler, Yunrong Guo, Ofir Nabati, Gal Chechik, and Xue~Bin Peng.
\newblock Maskedmimic: Unified physics-based character control through masked motion inpainting.
\newblock \emph{arXiv preprint arXiv:2409.14393}, 2024.

\bibitem[Tevet et~al.(2022)Tevet, Raab, Gordon, Shafir, Cohen-Or, and Bermano]{MDM}
Guy Tevet, Sigal Raab, Brian Gordon, Yonatan Shafir, Daniel Cohen-Or, and Amit~H Bermano.
\newblock Human motion diffusion model.
\newblock \emph{arXiv preprint arXiv:2209.14916}, 2022.

\bibitem[Thoduka et~al.(2024)Thoduka, Hochgeschwender, Gall, and Pl{\"o}ger]{thoduka2024multimodal}
Santosh Thoduka, Nico Hochgeschwender, Juergen Gall, and Paul~G Pl{\"o}ger.
\newblock A multimodal handover failure detection dataset and baselines.
\newblock \emph{arXiv preprint arXiv:2402.18319}, 2024.

\bibitem[Touvron et~al.(2023)Touvron, Martin, Stone, Albert, Almahairi, Babaei, Bashlykov, Batra, Bhargava, Bhosale, et~al.]{touvron2023llama}
Hugo Touvron, Louis Martin, Kevin Stone, Peter Albert, Amjad Almahairi, Yasmine Babaei, Nikolay Bashlykov, Soumya Batra, Prajjwal Bhargava, Shruti Bhosale, et~al.
\newblock Llama 2: Open foundation and fine-tuned chat models.
\newblock \emph{arXiv preprint arXiv:2307.09288}, 2023.

\bibitem[Unitree(2018)]{unitree_h1}
Unitree.
\newblock Unitree’s first universal humanoid robot.
\newblock \url{https://www.unitree.com/h1}, 2018.

\bibitem[Vaswani et~al.(2017)Vaswani, Shazeer, Parmar, Uszkoreit, Jones, Gomez, Kaiser, and Polosukhin]{vaswani2017attention}
Ashish Vaswani, Noam Shazeer, Niki Parmar, Jakob Uszkoreit, Llion Jones, Aidan~N Gomez, {\L}ukasz Kaiser, and Illia Polosukhin.
\newblock Attention is all you need.
\newblock \emph{Advances in neural information processing systems}, 30, 2017.

\bibitem[Wan et~al.(2022{\natexlab{a}})Wan, Yang, Liu, Zhang, Jia, Choi, Pan, Theobalt, Komura, and Wang]{HO-GCN}
Weilin Wan, Lei Yang, Lingjie Liu, Zhuoying Zhang, Ruixing Jia, Yi-King Choi, Jia Pan, Christian Theobalt, Taku Komura, and Wenping Wang.
\newblock Learn to predict how humans manipulate large-sized objects from interactive motions.
\newblock \emph{IEEE Robotics and Automation Letters}, 7\penalty0 (2):\penalty0 4702--4709, 2022{\natexlab{a}}.

\bibitem[Wan et~al.(2022{\natexlab{b}})Wan, Yang, Liu, Zhang, Jia, Choi, Pan, Theobalt, Komura, and Wang]{wan2022learn}
Weilin Wan, Lei Yang, Lingjie Liu, Zhuoying Zhang, Ruixing Jia, Yi-King Choi, Jia Pan, Christian Theobalt, Taku Komura, and Wenping Wang.
\newblock Learn to predict how humans manipulate large-sized objects from interactive motions.
\newblock \emph{IEEE Robotics and Automation Letters}, 7\penalty0 (2):\penalty0 4702--4709, 2022{\natexlab{b}}.

\bibitem[Wan et~al.(2023)Wan, Geng, Liu, Shan, Yang, Yi, and Wang]{wan2023unidexgrasp++}
Weikang Wan, Haoran Geng, Yun Liu, Zikang Shan, Yaodong Yang, Li Yi, and He Wang.
\newblock Unidexgrasp++: Improving dexterous grasping policy learning via geometry-aware curriculum and iterative generalist-specialist learning.
\newblock In \emph{Proceedings of the IEEE/CVF International Conference on Computer Vision}, pages 3891--3902, 2023.

\bibitem[Wang et~al.(2023)Wang, Lin, Zeng, Luo, Zhang, and Zhang]{wang2023physhoi}
Yinhuai Wang, Jing Lin, Ailing Zeng, Zhengyi Luo, Jian Zhang, and Lei Zhang.
\newblock Physhoi: Physics-based imitation of dynamic human-object interaction.
\newblock \emph{arXiv preprint arXiv:2312.04393}, 2023.

\bibitem[Wang et~al.(2020)Wang, Shin, and Fowlkes]{GPA}
Zhe Wang, Daeyun Shin, and Charless~C Fowlkes.
\newblock Predicting camera viewpoint improves cross-dataset generalization for 3d human pose estimation.
\newblock In \emph{Computer Vision--ECCV 2020 Workshops: Glasgow, UK, August 23--28, 2020, Proceedings, Part II 16}, pages 523--540. Springer, 2020.

\bibitem[Wang et~al.(2022)Wang, Chen, Liu, Zhu, Liang, and Huang]{HUMANISE}
Zan Wang, Yixin Chen, Tengyu Liu, Yixin Zhu, Wei Liang, and Siyuan Huang.
\newblock Humanise: Language-conditioned human motion generation in 3d scenes.
\newblock \emph{Advances in Neural Information Processing Systems}, 35:\penalty0 14959--14971, 2022.

\bibitem[Wang et~al.(2024)Wang, Chen, Jia, Li, Zhang, Zhang, Liu, Zhu, Liang, and Huang]{wang2024move}
Zan Wang, Yixin Chen, Baoxiong Jia, Puhao Li, Jinlu Zhang, Jingze Zhang, Tengyu Liu, Yixin Zhu, Wei Liang, and Siyuan Huang.
\newblock Move as you say interact as you can: Language-guided human motion generation with scene affordance.
\newblock In \emph{Proceedings of the IEEE/CVF Conference on Computer Vision and Pattern Recognition}, pages 433--444, 2024.

\bibitem[Wiederhold et~al.(2024)Wiederhold, Megyeri, Paris, Banerjee, and Banerjee]{HOH}
Noah Wiederhold, Ava Megyeri, DiMaggio Paris, Sean Banerjee, and Natasha Banerjee.
\newblock Hoh: Markerless multimodal human-object-human handover dataset with large object count.
\newblock \emph{Advances in Neural Information Processing Systems}, 36, 2024.

\bibitem[Wu et~al.(2024{\natexlab{a}})Wu, Shi, Huang, Yu, Xu, and Wang]{wu2024thor}
Qianyang Wu, Ye Shi, Xiaoshui Huang, Jingyi Yu, Lan Xu, and Jingya Wang.
\newblock Thor: Text to human-object interaction diffusion via relation intervention.
\newblock \emph{arXiv preprint arXiv:2403.11208}, 2024{\natexlab{a}}.

\bibitem[Wu et~al.(2023)Wu, Zhu, Peng, Hang, and Sun]{wu2023functional}
Rina Wu, Tianqiang Zhu, Wanli Peng, Jinglue Hang, and Yi Sun.
\newblock Functional grasp transfer across a category of objects from only one labeled instance.
\newblock \emph{IEEE Robotics and Automation Letters}, 8\penalty0 (5):\penalty0 2748--2755, 2023.

\bibitem[Wu et~al.(2022)Wu, Wang, Zhang, Zhang, Hilliges, Yu, and Tang]{SAGA}
Yan Wu, Jiahao Wang, Yan Zhang, Siwei Zhang, Otmar Hilliges, Fisher Yu, and Siyu Tang.
\newblock Saga: Stochastic whole-body grasping with contact.
\newblock In \emph{European Conference on Computer Vision}, pages 257--274. Springer, 2022.

\bibitem[Wu et~al.(2024{\natexlab{b}})Wu, Li, and Liu]{wu2024human}
Zhen Wu, Jiaman Li, and C~Karen Liu.
\newblock Human-object interaction from human-level instructions.
\newblock \emph{arXiv preprint arXiv:2406.17840}, 2024{\natexlab{b}}.

\bibitem[Xie et~al.(2023{\natexlab{a}})Xie, Bhatnagar, and Pons-Moll]{xie2023visibility}
Xianghui Xie, Bharat~Lal Bhatnagar, and Gerard Pons-Moll.
\newblock Visibility aware human-object interaction tracking from single rgb camera.
\newblock In \emph{Proceedings of the IEEE/CVF Conference on Computer Vision and Pattern Recognition}, pages 4757--4768, 2023{\natexlab{a}}.

\bibitem[Xie et~al.(2024{\natexlab{a}})Xie, Bhatnagar, Lenssen, and Pons-Moll]{xie2024template}
Xianghui Xie, Bharat~Lal Bhatnagar, Jan~Eric Lenssen, and Gerard Pons-Moll.
\newblock Template free reconstruction of human-object interaction with procedural interaction generation.
\newblock In \emph{Proceedings of the IEEE/CVF Conference on Computer Vision and Pattern Recognition}, pages 10003--10015, 2024{\natexlab{a}}.

\bibitem[Xie et~al.(2024{\natexlab{b}})Xie, Lenssen, and Pons-Moll]{xie2024intertrack}
Xianghui Xie, Jan~Eric Lenssen, and Gerard Pons-Moll.
\newblock Intertrack: Tracking human object interaction without object templates.
\newblock \emph{arXiv preprint arXiv:2408.13953}, 2024{\natexlab{b}}.

\bibitem[Xie et~al.(2023{\natexlab{b}})Xie, Tseng, Starke, van~de Panne, and Liu]{xie2023hierarchical}
Zhaoming Xie, Jonathan Tseng, Sebastian Starke, Michiel van~de Panne, and C~Karen Liu.
\newblock Hierarchical planning and control for box loco-manipulation.
\newblock \emph{Proceedings of the ACM on Computer Graphics and Interactive Techniques}, 6\penalty0 (3):\penalty0 1--18, 2023{\natexlab{b}}.

\bibitem[Xu et~al.(2023{\natexlab{a}})Xu, Song, Wang, Su, Fang, Ding, Gan, Yan, Jin, Yang, et~al.]{xu2023actformer}
Liang Xu, Ziyang Song, Dongliang Wang, Jing Su, Zhicheng Fang, Chenjing Ding, Weihao Gan, Yichao Yan, Xin Jin, Xiaokang Yang, et~al.
\newblock Actformer: A gan-based transformer towards general action-conditioned 3d human motion generation.
\newblock In \emph{Proceedings of the IEEE/CVF International Conference on Computer Vision}, pages 2228--2238, 2023{\natexlab{a}}.

\bibitem[Xu et~al.(2023{\natexlab{b}})Xu, Li, Wang, and Gui]{InterDiff}
Sirui Xu, Zhengyuan Li, Yu-Xiong Wang, and Liang-Yan Gui.
\newblock Interdiff: Generating 3d human-object interactions with physics-informed diffusion.
\newblock In \emph{Proceedings of the IEEE/CVF International Conference on Computer Vision}, pages 14928--14940, 2023{\natexlab{b}}.

\bibitem[Xu et~al.(2024)Xu, Wang, Wang, and Gui]{InterDreamer}
Sirui Xu, Ziyin Wang, Yu-Xiong Wang, and Liang-Yan Gui.
\newblock Interdreamer: Zero-shot text to 3d dynamic human-object interaction.
\newblock \emph{arXiv preprint arXiv:2403.19652}, 2024.

\bibitem[Xu et~al.(2023{\natexlab{c}})Xu, Wan, Zhang, Liu, Shan, Shen, Wang, Geng, Weng, Chen, et~al.]{xu2023unidexgrasp}
Yinzhen Xu, Weikang Wan, Jialiang Zhang, Haoran Liu, Zikang Shan, Hao Shen, Ruicheng Wang, Haoran Geng, Yijia Weng, Jiayi Chen, et~al.
\newblock Unidexgrasp: Universal robotic dexterous grasping via learning diverse proposal generation and goal-conditioned policy.
\newblock In \emph{Proceedings of the IEEE/CVF Conference on Computer Vision and Pattern Recognition}, pages 4737--4746, 2023{\natexlab{c}}.

\bibitem[Yan et~al.(2024{\natexlab{a}})Yan, Cui, Xie, and Guo]{yan2024forecasting}
Haitao Yan, Qiongjie Cui, Jiexin Xie, and Shijie Guo.
\newblock Forecasting of 3d whole-body human poses with grasping objects.
\newblock In \emph{Proceedings of the IEEE/CVF Conference on Computer Vision and Pattern Recognition}, pages 1726--1736, 2024{\natexlab{a}}.

\bibitem[Yan et~al.(2023)Yan, Wang, Dai, Shen, Wen, Xu, Ma, and Wang]{CIMI4D}
Ming Yan, Xin Wang, Yudi Dai, Siqi Shen, Chenglu Wen, Lan Xu, Yuexin Ma, and Cheng Wang.
\newblock Cimi4d: A large multimodal climbing motion dataset under human-scene interactions.
\newblock In \emph{Proceedings of the IEEE/CVF Conference on Computer Vision and Pattern Recognition}, pages 12977--12988, 2023.

\bibitem[Yan et~al.(2024{\natexlab{b}})Yan, Zhang, Cai, Fan, Lin, Dai, Shen, Wen, Xu, Ma, et~al.]{yan2024reli11d}
Ming Yan, Yan Zhang, Shuqiang Cai, Shuqi Fan, Xincheng Lin, Yudi Dai, Siqi Shen, Chenglu Wen, Lan Xu, Yuexin Ma, et~al.
\newblock Reli11d: A comprehensive multimodal human motion dataset and method.
\newblock \emph{arXiv preprint arXiv:2403.19501}, 2024{\natexlab{b}}.

\bibitem[Yan et~al.(2019)Yan, Li, Xiong, Yan, and Lin]{yan2019convolutional}
Sijie Yan, Zhizhong Li, Yuanjun Xiong, Huahan Yan, and Dahua Lin.
\newblock Convolutional sequence generation for skeleton-based action synthesis.
\newblock In \emph{Proceedings of the IEEE/CVF International Conference on Computer Vision}, pages 4394--4402, 2019.

\bibitem[Yang et~al.(2024)Yang, Niu, Jiang, Zhang, and Huang]{F-HOI}
Jie Yang, Xuesong Niu, Nan Jiang, Ruimao Zhang, and Siyuan Huang.
\newblock F-hoi: Toward fine-grained semantic-aligned 3d human-object interactions.
\newblock \emph{arXiv preprint arXiv:2407.12435}, 2024.

\bibitem[Yang et~al.(2021)Yang, Zhan, Li, Xu, Li, and Lu]{yang2021cpf}
Lixin Yang, Xinyu Zhan, Kailin Li, Wenqiang Xu, Jiefeng Li, and Cewu Lu.
\newblock Cpf: Learning a contact potential field to model the hand-object interaction.
\newblock In \emph{Proceedings of the IEEE/CVF International Conference on Computer Vision}, pages 11097--11106, 2021.

\bibitem[Yang et~al.(2022)Yang, Li, Zhan, Wu, Xu, Liu, and Lu]{yang2022oakink}
Lixin Yang, Kailin Li, Xinyu Zhan, Fei Wu, Anran Xu, Liu Liu, and Cewu Lu.
\newblock Oakink: A large-scale knowledge repository for understanding hand-object interaction.
\newblock In \emph{Proceedings of the IEEE/CVF Conference on Computer Vision and Pattern Recognition}, pages 20953--20962, 2022.

\bibitem[Ye et~al.(2021)Ye, Xu, Xue, Tang, Wang, and Lu]{ye2021h2o}
Ruolin Ye, Wenqiang Xu, Zhendong Xue, Tutian Tang, Yanfeng Wang, and Cewu Lu.
\newblock H2o: A benchmark for visual human-human object handover analysis.
\newblock In \emph{Proceedings of the IEEE/CVF International Conference on Computer Vision}, pages 15762--15771, 2021.

\bibitem[Yi et~al.(2024)Yi, Thies, Black, Peng, and Rempe]{yi2024generating}
Hongwei Yi, Justus Thies, Michael~J Black, Xue~Bin Peng, and Davis Rempe.
\newblock Generating human interaction motions in scenes with text control.
\newblock \emph{arXiv preprint arXiv:2404.10685}, 2024.

\bibitem[Yi et~al.(2025)Yi, Thies, Black, Peng, and Rempe]{yi2025generating}
Hongwei Yi, Justus Thies, Michael~J Black, Xue~Bin Peng, and Davis Rempe.
\newblock Generating human interaction motions in scenes with text control.
\newblock In \emph{European Conference on Computer Vision}, pages 246--263. Springer, 2025.

\bibitem[Zhan et~al.(2024)Zhan, Yang, Zhao, Mao, Xu, Lin, Li, and Lu]{OAKINK2}
Xinyu Zhan, Lixin Yang, Yifei Zhao, Kangrui Mao, Hanlin Xu, Zenan Lin, Kailin Li, and Cewu Lu.
\newblock Oakink2: A dataset of bimanual hands-object manipulation in complex task completion.
\newblock In \emph{Proceedings of the IEEE/CVF Conference on Computer Vision and Pattern Recognition}, pages 445--456, 2024.

\bibitem[Zhang et~al.(2024{\natexlab{a}})Zhang, Xiao, He, and Shi]{zhang2024wococo}
Chong Zhang, Wenli Xiao, Tairan He, and Guanya Shi.
\newblock Wococo: Learning whole-body humanoid control with sequential contacts.
\newblock \emph{arXiv preprint arXiv:2406.06005}, 2024{\natexlab{a}}.

\bibitem[Zhang et~al.(2024{\natexlab{b}})Zhang, Christen, Fan, Zheng, Hwangbo, Song, and Hilliges]{zhang2024artigrasp}
Hui Zhang, Sammy Christen, Zicong Fan, Luocheng Zheng, Jemin Hwangbo, Jie Song, and Otmar Hilliges.
\newblock Artigrasp: Physically plausible synthesis of bi-manual dexterous grasping and articulation.
\newblock In \emph{2024 International Conference on 3D Vision (3DV)}, pages 235--246. IEEE, 2024{\natexlab{b}}.

\bibitem[Zhang et~al.(2023{\natexlab{a}})Zhang, Luo, Yang, Xu, Wu, Shi, Yu, Xu, and Wang]{NeuralDome}
Juze Zhang, Haimin Luo, Hongdi Yang, Xinru Xu, Qianyang Wu, Ye Shi, Jingyi Yu, Lan Xu, and Jingya Wang.
\newblock Neuraldome: A neural modeling pipeline on multi-view human-object interactions.
\newblock In \emph{Proceedings of the IEEE/CVF Conference on Computer Vision and Pattern Recognition}, pages 8834--8845, 2023{\natexlab{a}}.

\bibitem[Zhang et~al.(2024{\natexlab{c}})Zhang, Zhang, Song, Shi, Zhao, Shi, Yu, Xu, and Wang]{HOI-M3}
Juze Zhang, Jingyan Zhang, Zining Song, Zhanhe Shi, Chengfeng Zhao, Ye Shi, Jingyi Yu, Lan Xu, and Jingya Wang.
\newblock Hoi-m3: Capture multiple humans and objects interaction within contextual environment.
\newblock \emph{arXiv preprint arXiv:2404.00299}, 2024{\natexlab{c}}.

\bibitem[Zhang et~al.(2022{\natexlab{a}})Zhang, Ma, Zhang, Qian, Kwon, Pollefeys, Bogo, and Tang]{EgoBody}
Siwei Zhang, Qianli Ma, Yan Zhang, Zhiyin Qian, Taein Kwon, Marc Pollefeys, Federica Bogo, and Siyu Tang.
\newblock Egobody: Human body shape and motion of interacting people from head-mounted devices.
\newblock In \emph{European Conference on Computer Vision}, pages 180--200. Springer, 2022{\natexlab{a}}.

\bibitem[Zhang et~al.(2023{\natexlab{b}})Zhang, Dabral, Leimk{\"u}hler, Golyanik, Habermann, and Theobalt]{ROAM}
Wanyue Zhang, Rishabh Dabral, Thomas Leimk{\"u}hler, Vladislav Golyanik, Marc Habermann, and Christian Theobalt.
\newblock Roam: Robust and object-aware motion generation using neural pose descriptors.
\newblock \emph{arXiv preprint arXiv:2308.12969}, 1, 2023{\natexlab{b}}.

\bibitem[Zhang et~al.(2022{\natexlab{b}})Zhang, Bhatnagar, Starke, Guzov, and Pons-Moll]{COUCH}
Xiaohan Zhang, Bharat~Lal Bhatnagar, Sebastian Starke, Vladimir Guzov, and Gerard Pons-Moll.
\newblock Couch: Towards controllable human-chair interactions.
\newblock In \emph{European Conference on Computer Vision}, pages 518--535. Springer, 2022{\natexlab{b}}.

\bibitem[Zhang et~al.(2024{\natexlab{d}})Zhang, Bhatnagar, Starke, Petrov, Guzov, Dhamo, P{\'e}rez-Pellitero, and Pons-Moll]{FORCE}
Xiaohan Zhang, Bharat~Lal Bhatnagar, Sebastian Starke, Ilya Petrov, Vladimir Guzov, Helisa Dhamo, Eduardo P{\'e}rez-Pellitero, and Gerard Pons-Moll.
\newblock Force: Dataset and method for intuitive physics guided human-object interaction.
\newblock \emph{arXiv preprint arXiv:2403.11237}, 2024{\natexlab{d}}.

\bibitem[Zhao et~al.(2024)Zhao, Zhang, Du, Shan, Wang, Yu, Wang, and Xu]{zhao2024m}
Chengfeng Zhao, Juze Zhang, Jiashen Du, Ziwei Shan, Junye Wang, Jingyi Yu, Jingya Wang, and Lan Xu.
\newblock I'm hoi: Inertia-aware monocular capture of 3d human-object interactions.
\newblock In \emph{Proceedings of the IEEE/CVF Conference on Computer Vision and Pattern Recognition}, pages 729--741, 2024.

\bibitem[Zhao et~al.(2022)Zhao, Wang, Zhang, Beeler, and Tang]{COINS}
Kaifeng Zhao, Shaofei Wang, Yan Zhang, Thabo Beeler, and Siyu Tang.
\newblock Compositional human-scene interaction synthesis with semantic control.
\newblock In \emph{European Conference on Computer Vision}, pages 311--327. Springer, 2022.

\bibitem[Zhao et~al.(2023)Zhao, Kumar, Levine, and Finn]{ACT}
Tony~Z Zhao, Vikash Kumar, Sergey Levine, and Chelsea Finn.
\newblock Learning fine-grained bimanual manipulation with low-cost hardware.
\newblock \emph{arXiv preprint arXiv:2304.13705}, 2023.

\bibitem[Zheng et~al.(2023)Zheng, Zheng, Fang, Liu, and Yi]{zheng2023cams}
Juntian Zheng, Qingyuan Zheng, Lixing Fang, Yun Liu, and Li Yi.
\newblock Cams: Canonicalized manipulation spaces for category-level functional hand-object manipulation synthesis.
\newblock In \emph{Proceedings of the IEEE/CVF Conference on Computer Vision and Pattern Recognition}, pages 585--594, 2023.

\bibitem[Zhu et~al.(2024)Zhu, Li, and Jakab]{zhu2024dreamhoi}
Thomas~Hanwen Zhu, Ruining Li, and Tomas Jakab.
\newblock Dreamhoi: Subject-driven generation of 3d human-object interactions with diffusion priors.
\newblock \emph{arXiv preprint arXiv:2409.08278}, 2024.

\end{thebibliography}
}

\clearpage
\Large \textbf{Appendix}
\\
\normalsize
\appendix
{\Large \textbf{Contents:}}

\begin{itemize}
\item \ref{supp_sec:cross_dataset_evaluation}. Cross-dataset Evaluation
\item \ref{supp_sec:details_real_world_data_aquisition}. Details on Real-world Data Acquisition
\item \ref{supp_sec:details_collaboration_retargeting}. Details on \dataset-Synthetic Data Generation
\item \ref{supp_sec:dataset_statistics}. Dataset Statistics and Visualization
\item \ref{supp_sec:data_split_details}. Details on Data Split
\item \ref{supp_sec:metrics_benchmarks}. Evaluation Metrics for Benchmarks
\item \ref{supp_sec:qualitative_results_on_benchmarks}. Qualitative Results on Benchmarks
\item \ref{supp_sec:application_details}. Details on the Application of \dataset-Synthetic
\item \ref{supp_sec:details_on_humanoid_skill_learning}. Details on Humanoid Skill Learning using \dataset
\item \ref{supp_sec:data_capture_instructions_and_costs}. \dataset-Real Data Capturing Instructions
\end{itemize}

\section{Cross-dataset Evaluation}
\label{supp_sec:cross_dataset_evaluation}

To examine the data quality of \dataset-Real, we follow existing dataset efforts~\cite{CIMI4D,chao2021dexycb,liu2024taco} and conduct the vision-based cross-dataset evaluation. We select an individual human-object-interaction dataset BEHAVE~\cite{BEHAVE} that includes color images and select 2D human keypoint estimation as the evaluation task.

\textbf{Data Preparation.} For a color image from \dataset-Real and BEHAVE~\cite{BEHAVE}, we first detect the bounding box for each person via ground truth human pose and obtain the image patch for the person. We then resize the image patch to get a maximal length of 256 pixels and fill it up into a 256x256 image with the black color as the background. Finally, for each 256x256 image, we automatically acquire the ground truth 2D-pixel coordinates of 22 SMPL-X~\cite{SMPLX} human body joints from 3D human poses. For data split, we follow the original train-test split for BEHAVE~\cite{BEHAVE} and merge the two test sets (S1, S2) for \dataset-Real.

\textbf{Task Formulation.} Given a 256x256 color image including a person, the task is to estimate the 2D-pixel coordinate for each of the 22 SMPL-X~\cite{SMPLX} human body joints.

\textbf{Evaluation Metrics.} $P_e$ denotes the mean-square error of 2D coordinate estimates.
$Acc$ denotes the percentage of the coordinate estimates with the Euclidean distance to the ground truth smaller than $15$ pixels.

\textbf{Method, Results, and Analysis.} We draw inspiration from HybrIK-X~\cite{li2023hybrik} and adopt their vision backbone as the solution. Table~\ref{tab:cross_dataset_validation} shows the method performances on the two datasets under different training settings. Due to the significant domain gaps in visual patterns and human behaviors, transferring models trained on one dataset to the other would consistently encounter error increases. Despite the domain gaps, integrally training on both datasets achieves large performance gains on both \dataset-Real and BEHAVE~\cite{BEHAVE}, indicating the accuracy of \dataset-Real and the value of the dataset serving for visual perception studies.

\begin{table}[h]
\centering
\footnotesize
{
\begin{tabular}{|c|c|c|c|}
\hline  
\backslashbox{Test}{Train} & \dataset-Real & BEHAVE~\cite{BEHAVE} & \makecell{\dataset-Real\\+BEHAVE~\cite{BEHAVE}} \\
\hline
\dataset-Real & 152.4 / 91.2 & 904.9 / 35.6 & \textbf{121.7} / \textbf{92.4} \\
\hline
BEHAVE\cite{BEHAVE} & 887.9 / 37.8 & 146.3 / 88.9 & \textbf{128.2} / \textbf{89.8} \\
\hline
\end{tabular}
\caption{\textbf{Cross-dataset evaluation with BEHAVE~\cite{BEHAVE} on 2D human keypoint estimation.} Results are in $P_e$ (pixel$^2$, lower is better) and $Acc$ ($\%$, higher is better), respectively.}
\label{tab:cross_dataset_validation}
}
\end{table}




\section{Details on Real-world Data Aquisition}
\label{supp_sec:details_real_world_data_aquisition}

In this section, we describe our system calibration (Section \ref{sec:sys_calib}) and time synchronization (Section \ref{sec:time_sync}) in detail. Moreover, we provide detailed information on loss functions of the human mesh acquisition (Section \ref{sec:loss_human}).

\subsection{System Calibration}
\label{sec:sys_calib}

\textbf{Calibrating the Inertial-optical Mocap System.} Three reflective markers are fixed at known positions on a calibration rod, by which the 12 high-speed motion capture cameras calculate their relative extrinsic parameters, providing information about their spatial relationships. Additionally, three markers fixed at the world coordinate origin are employed to calibrate the motion capture system coordinate with the defined world coordinate.

\textbf{Calibrating Camera Intrinsic.} The intrinsic parameters of allocentric and egocentric cameras are calibrated using a chessboard pattern.

\textbf{Calibrating Extrinsic of the Allocentric Cameras.} We place ten markers in the camera view to locate each allocentric camera. By annotating the markers' 3D positions in the world coordinate system and their 2D-pixel coordinates on allocentric images, the camera's extrinsic parameters are estimated by solving a Perspective-n-Point (PnP) problem via OpenCV.

\textbf{Calibrating Extrinsic of the Egocentric Camera.} We obtain the camera's pose information by fixing the camera to the head tracker of the motion capture suit. Similarly, ten markers are used to calibrate the relative extrinsic parameters of the first-person perspective cameras, allowing for determining their positions and orientations relative to the motion capture system. Additionally, to mitigate errors introduced by the integration of optical and inertial tracking systems, a purely optical tracking rigid is mounted on the motion camera.

\subsection{Time Synchronization} 
\label{sec:time_sync}
To implement our synchronization method, we first set up a Network Time Protocol (NTP) server on the motion capture host. This server serves as the time synchronization reference for the Windows computer connected to the Kinect Azure DK. We minimize time discrepancies by connecting the Windows computer to the NTP server in high-precision mode and thus achieving precise synchronization.

Additionally, we employ a Linear Timecode (LTC) generator to encode a time signal onto the action camera's audio track. This time signal serves as a synchronization reference for aligning the first-person perspective RGB information with the motion capture data.

\subsection{Loss Function Designs for Human Mesh Acquisition}
\label{sec:loss_human}
To transfer the BVH~\cite{meredith2001motion} human skeleton to the widely-used SMPL-X~\cite{SMPLX} model. We optimize body shape parameters $\beta \in \mathbb{R}^{10}$ to fit the constraints on manually measured human skeleton lengths and then optimize the full-body pose $\theta \in \mathbb{R}^{159}$ with the following loss function:
\begin{align}
    \mathcal{L} = \mathcal{L}_{\text{reg}} + \mathcal{L}_{j\text{3D}} + \mathcal{L}_{j\text{Ori}} + \mathcal{L}_{\text{smooth}} + \mathcal{L}_{h\text{3D}} + \mathcal{L}_{h\text{Ori}} + \mathcal{L}_{\text{contact}}.
\end{align}

\textbf{Regularization Loss $\mathcal{L}_{\text{reg}}$}. The regularization loss term is defined as
\begin{align}
    \mathcal{L}_{\text{reg}} = \sum\left|\left|\theta_{\text{body}}\right|\right|^2 \cdot \lambda_{\text{body}} + \left( \sum\left|\left|\theta_{l\_\text{hand}}\right|\right|^2 + \sum\left|\left|\theta_{r\_\text{hand}}\right|\right|^2 \right) \cdot \lambda_{\text{hand}},
\end{align}
where $\theta_{\text{body}} \in \mathbb{R}^{21\times3}$ represents the body pose parameters defined by 21 joints of the skeleton, $\theta_{l\_hand} \in \mathbb{R}^{12}$ and $\theta_{r\_\text{hand}} \in \mathbb{R}^{12}$ represents the hand pose parameters. For each hand, the original SMPL-X skeleton has 15 joints with parameters $\theta_{\text{hand}} \in \mathbb{R}^{15\times3}$. However, principal component analysis (PCA) is applied to the hand pose parameters. The $\theta_{\text{hand}}$ parameters are transformed into a lower-dimensional space, specifically $\mathbb{R}^{12}$. $\lambda_{\text{body}}=10^{-3}$ and $\lambda_{\text{hand}}=10^{-4}$ are different weights that are used to control the regularization strength for the body and hand pose parameters, respectively. This loss ensures the simplicity of the results and prevents unnatural, significant twisting of the joints.

\textbf{3D Position Loss $\mathcal{L}_{j\text{3D}}$ and $\mathcal{L}_{h\text{3D}}$}. The 3D position loss term is defined as
\begin{align}
    \mathcal{L}_{\text{3D}} = \sum\left|\left| \textbf{T}_{\text{smplx}} - \textbf{T}_{\text{bvh}}\right|\right|^2 \cdot \lambda_{\text{3D}},
\end{align}
where $\textbf{T}_{\text{smplx}} \in \mathbb{R}^{3}$ represents the 3D global coordinates of the joints in the SMPL-X model and $\textbf{T}_{\text{bvh}} \in \mathbb{R}^{3}$ represents the corresponding 3D global coordinates of the joints in the BVH representation. $\mathcal{L}_{j\text{3D}}$ represents the 3D position loss sum for the 21 body joints, while $\mathcal{L}_{h\text{3D}}$ represents the 3D position loss sum for the 30 hand joints (15 joints per hand). These two terms have different weights, set as $\lambda_{j\text{3D}}=1.0$ and $\lambda_{h\text{3D}}=2.0$, respectively.

\textbf{Orientation Loss $\mathcal{L}_{j\text{Ori}}$ and $\mathcal{L}_{h\text{Ori}}$}. The orientation loss term is defined as
\begin{align}
    \mathcal{L}_{\text{Ori}} = \sum\left|\left| \textbf{R}_{\text{smplx}} - \textbf{R}_{\text{bvh}}\right|\right|^2 \cdot \lambda_{\text{Ori}},
\end{align}
which is similar to $\mathcal{L}_{\text{3D}}$, except that $\mathcal{R}_{\text{smplx} }\in \mathbb{R}^{3\times3}$ and $\mathcal{R}_{\text{bvh}} \in \mathbb{R}^{3\times3}$ represent the rotation matrices for the adjacent joints in the SMPL-X and corresponding BVH representations, respectively. Specifically, body joints named head, spine, spine2, leftUpLeg, rightUpLeg, rightShoulder, leftShoulder, rightArm, leftArm, and neck are subjected to orientation loss, ensuring that their rotations relative to adjacent nodes are close to the BVH ground truth. $\lambda_{\text{Ori}}$ is set to $0.2$.

\textbf{Temporal Smoothness Loss $\mathcal{L}_{\text{smooth}}$}. The temporal smoothness loss term is defined as
\begin{align}
    \mathcal{L}_{\text{smooth}} = \sum_{i=1}^{N} \left( \left| \left| \theta_{i} -  \theta_{i-1}\right|\right|^2 \right) \cdot \lambda_{\text{smooth}}
\end{align}
where $\theta_{i} \in \mathbb{R}^{(21+30)\times3}$ represents the body and hand pose of the $i$-th frame. $\lambda_{\text{smooth}}$ is set to $20.0$.

\textbf{Contact Loss $\mathcal{L}_{\text{contact}}$}. The contact loss term is defined as
\begin{align}
    \mathcal{L}_{\text{contact}} = \sum \left( \left| \left| \textbf{T}_{\text{finger}} -  \textbf{T}_{\text{obj}}\right|\right|^2 \cdot \mathcal{J}(\textbf{T}_{\text{finger}}, \textbf{T}_{\text{obj}}) \right) \cdot \lambda_{\text{contact}}
\end{align}
where $\mathcal{T}_{\text{finger}} \in \mathbb{R}^{10\times3}$ is the global coordinates of ten fingers, and $\mathcal{T}_{\text{obj}} \in \mathbb{R}^{10\times3}$ is the corresponding global coordinates of the point closest to finger. $\mathcal{J}(\textbf{T}_{\text{finger}}, \textbf{T}_{\text{obj}})$ is 1 when the distance between $\textbf{T}_{\text{finger}}$ and $\textbf{T}_{\text{obj}}$ is less than a threshold, otherwise it is 0. And $\lambda_{\text{contact}}$ is $2.0$.



\section{Details on \dataset-Synthetic Data Generation}
\label{supp_sec:details_collaboration_retargeting}



In this section, we provide details on our synthetic data generation (collaboration retargeting) method. Firstly, we clarify term definitions in Section \ref{supp_sec:term_definition}. We then explicitly introduce the whole method pipeline in detail in Section \ref{supp_sec:method_pipeline}. Finally, we provide implementation details in Sections \ref{supp_sec:contact-guided_interaction_retargeting} and \ref{supp_sec:human_pose_discriminator}.

\subsection{Term Definitions}
\label{supp_sec:term_definition}
We provide definitions for the terms in our collaboration retargeting pipeline as follows.

\textbf{Contact Candidate}: Contact candidate is a quadruple list containing all possible contact region index (person1\_leftHand, person1\_rightHand, person2\_leftHand, person2\_rightHand) on \textit{source}'s vertices. For each \textit{source}, we record the contact regions of the four hands in each frame of each data sequence. At the beginning of the synthetic data generation pipeline, we sample contact candidates from these records.

\textbf{Contact Constraint}: Having contact candidate on \textit{source}, we apply DeepSDF-based~\cite{park2019deepsdf} contact retargeting to transfer the contact regions to \textit{target}. These contact regions on \textit{target} are the contact constraints fed into the contact-guided interaction retargeting module.

\textbf{Source Interaction}:
During each collaboration retargeting process, we sample a human-object-human collaborative motion sequence from \dataset-Real as the source interaction to guide temporal collaboration pattern.

\textbf{Interaction Candidate}: Sampling $N$ contact candidates, we apply contact-guided interaction retargeting $N$ times and have $N$ human-object-human motion outputs, dubbed interaction candidates. These motions would be fed into the human-centric contact selection module to assess their naturalness.

\subsection{Method Pipeline}
\label{supp_sec:method_pipeline}

The algorithm takes a \textit{source}-\textit{target} pair as input. First, we sample contact candidates from the whole \dataset-Real contact knowledge on \textit{source}. For each contact candidate, we apply object-centric contact retargeting to propagate contact candidates to contact constraints on \textit{target}. Sampling motion from \dataset-Real provides a high-level temporal collaboration pattern, and together with augmented low-level spatial relations, we obtain interaction candidates from the contact-guided interaction retargeting. Then, the human-centric contact selection module selects the optimal candidates, prompting a contact constraint update. After multiple iterations, the process yields augmented interactions. This iterative mechanism ensures a refined augmentation of interactions, enhancing the dataset's applicability across various scenarios.

\subsection{Contact-guided Interaction Retargeting}
\label{supp_sec:contact-guided_interaction_retargeting}
The contact-guided interaction retargeting is a two-step optimization. We start by optimizing the motion of \textit{target}. Then with \textit{target} contact constraints, we optimize the poses of the two persons.

\textbf{Object motion retargeting.} We deliberately design temporal and spatial losses to acquire consistent and smooth \textit{target} motion. In the concern of efficiency, we jointly optimize all frames in a single data sequence with $N$ frames. To guarantee the fidelity of object motion, we design the fidelity loss $L_f$ to restrict the rotation $R_{o,i}$ and the translation $T_{o,i}$ with the ground-truth rotation $R'_{o,i}$ and translation $T'_{o,i}$ in $N$ frames: 
\begin{align}
    \mathcal{L}_{f} = \lambda_{f} \sum\limits_{i}(||R'_{o,i} - R_{o,i}||_{1} + ||T'_{o,i} - T_{o,i}||_{1}).
\end{align}
We then address restriction on \textit{target}'s spatial position to avoid penetration with the ground. The spatial loss is defined as:


\begin{align}
    \mathcal{L}_{\text{spat}} = \lambda_{\text{spat}} \sum\limits_{i} \text{max} (- \text{min}(\text{height}_{i}), 0),
    \label{eq:L_spat}
\end{align}

where $\text{min}(\text{height}_{i})$ represents the lowest spatial position of the objects per frame.
A smoothness loss is designed to constrain the object pose difference between consecutive frames:
\begin{align}
    \mathcal{L}_{\text{smooth}} = \lambda_{\text{smooth}} \sum\limits_{i}a_{R_{o,i}}^2 + a_{T_{o,i}}^2,
    \label{eq:L_smooth}
\end{align}
where $a$ is the acceleration of rotation and translation during $N$ frames defined as:
\begin{align}
    a_{R_{o,i}} = 2R_{o,i} - R_{o,i-1} - R_{o,i+1}, \\
    a_{T_{o,i}} = 2T_{o,i} - T_{o,i-1} - T_{o,i+1},
\end{align}
The total object motion retargeting problem is:
\begin{align}
    R_o, T_o \longleftarrow{} \mathop{\operatorname{argmin}}\limits_{R_o, T_o}(\mathcal{L}_{f} + \mathcal{L}_{\text{spat}} + \mathcal{L}_{\text{smooth}}).
\end{align}

\textbf{Human motion retargeting.} We next optimize each person's motion based on the motion of \textit{target} and the contact constraint. To acquire visually plausible motion, we design the fidelity loss $\mathcal{L}_{j}$ and the smoothness loss $\mathcal{L}_{\text{smooth}}$. Besides, we utilize the contact correctness loss $\mathcal{L}_{c}$ to acquire contact consistency in \textit{target} interaction motion, and leverage spatial loss $L_\text{spat}$ similar to Equation~\ref{eq:L_spat} to avoid human-ground inter-penetration.

To enhance motion fidelity, we define two loss functions $\mathcal{L}_{\text{sr}}$ and $\mathcal{L}_{\text{wr}}$ and let $L_j = \mathcal{L}_{\text{sr}} + \mathcal{L}_{\text{wr}}$. For joints from the human arms, despite following the correct temporal collaboration pattern, their global positions would vary concerning diverse object geometries. Therefore, we utilize oriented vectors pointing to their parent body joints to obtain a relative joint fidelity:

\begin{align}
    \mathcal L_{\text{sr}} = \lambda_{\text{sr}} \sum\limits_{i} \sum\limits_{j \in \text{arm}} \Vert (P_{j,i} - P_{\text{parent}(j),i}) - (P'_{j,i} - P'_{\text{parent}(j),i}) \Vert_2^2,
\end{align}

where $P_{j,i}$ denotes the 3D global position of joint $j$ in frame $i$, and $P'$ denotes ground-truth values. $\mathcal{L}_{\text{wr}}$ denotes constraints on the global positions of other joints:

\begin{align}
    \mathcal L_{\text{wr}} = \lambda_{\text{wr}} \sum\limits_{i} \sum\limits_{j \notin \text{arm}} \Vert P_{j,i} - P'_{j,i} \Vert_2^2.
\end{align}


The design of the smoothness loss is similar to Equation~\ref{eq:L_smooth}, penalizing huge acceleration of human SMPL-X parameters to avoid great motion differences between frames:
\begin{align}
    \mathcal{L}_{\text{smooth}} = \lambda_{\text{smooth}} \sum\limits_{i}\sum\limits_{j\in \{1,2\}}(a_{\theta_{j,i}})^2 + (a_{T_{j,i}})^2 + (a_{O_{j,i}})^2.
\end{align}

To leverage contact constraints, we attract human hands to the corresponding contact region on \textit{target}. We select the positions of 20 fingertips of the two persons in the $i$-th frame as $\mathcal{H}_i = \{\bar{P}_{\text{tip,i}}\}_{\text{tip}\in[1,20]}$, where $\bar{P}$ are tip positions in the object's coordinate system. The contact vertices on the \textit{target} from object-centric contact retargeting are defined as $\mathcal{C} = \{\bar{P}'_{\text{tip}}\}_{\text{tip}\in[1,20]}$. We minimize the Chamfer Distance ($CD$) between $\mathcal{H}_i$ and $\mathcal{C}$ to obtain contact consistency:
\begin{align}
    \mathcal{L}_{c} = \lambda_c \sum\limits_{i} CD(\mathcal{H}_i, \mathcal{C}).
\end{align}
The total human motion retargeting problem is:

\begin{align}
    \theta_{1,2}, T_{1,2}, O_{1,2} \longleftarrow \mathop{\operatorname{argmin}}\limits_{\theta_{1,2}, T_{1,2}, O_{1,2}}(\mathcal{L}_{j} + \mathcal{L}_{c} + \mathcal{L}_{\text{spat}} + \mathcal{L}_{\text{smooth}}),
\end{align}

In practice, we run 1,000 and 1,500 iterations respectively for object motion retargeting and human motion retargeting. The whole pipeline is implemented in PyTorch with Adam solver. The learning rate is 0.01. In object motion retargeting, $\lambda_f$ for rotation is 500, for translation is 0.005, $\lambda_{\text{spat}}=0.01$, $\lambda_{\text{smooth}}=1$. In human motion retargeting, $\lambda_{\text{sr}} = 0.1$, $\lambda_{\text{wr}} = 0.003$, $\lambda_c = 1,000$, $\lambda_{\text{spat}}=0.01$, and $\lambda_{\text{smooth}} = 1$.
\subsection{Human-centric contact selection}
\label{supp_sec:human_pose_discriminator}

The pairwise training dataset utilized for the human pose discriminator training comprises 636,424 pairs of data. Each pair encompasses a positive human pose $S_{\text{pos}} \in \mathbb{R}^{21\times3}$ and a negative human pose $S_{\text{neg}} \in \mathbb{R}^{21\times3}$. The positive human pose is sampled from the \dataset-Real. Conversely, the negative human pose is derived from the corresponding positive sample by introducing noise to its object pose, subsequently employing the original contact information to perform contact-guided interaction retargeting. The discriminator is trained by:
\begin{align}
\mathcal{L}_{\text{ranking}} = - \log(\sigma(R_{\text{pos}} - R_{\text{neg}} - m(S_{\text{pos}}, S_{\text{neg}}))),
\end{align}
iterating 1,000 epochs by the Adam solver with a learning rate 2e-4.

Specifically, the noise $\Delta(\alpha, \beta, \gamma, x, y, z)$ incorporates both rotational and translational components. The rotational noise $\Delta(\alpha, \beta, \gamma)$ ranges from 20 to 60 degrees, while the translational noise $\Delta(x, y, z)$ falls within the range of 0.2 to 0.5 meters. The margin is computed by:
\begin{align}
    m(S_{\text{pos}}, S_{\text{neg}}) = (|\alpha| + |\beta| + |\gamma|) / 10 + (|x| + |y| + |z|) *10. 
\end{align}

During the contact constraint update process, a penetration filtering step is performed. For each frame, the penetration volume between the human and object is calculated. If the penetration volume exceeds $10^{-4}$ cubic meters, it is considered a penetration case. If more than 2.5\% of frames within an interaction candidate exhibit penetration, the entire candidate is discarded. Among the remaining candidates, the one with the highest score from the human pose discriminator is selected to proceed with the contact constraint update.


\section{Dataset Statistics and Visualization}
\label{supp_sec:dataset_statistics}


\begin{table*}[tb]
  \centering
  \footnotesize
  \begin{tabular}{|c|cccccc|cccccc|}
    \hline
    \multirow{2}{*}{Set} & \multicolumn{6}{c|}{$\#$Object} & \multicolumn{6}{c|}{$\#$Sequence} \\
    \cline{2-13}
    & Chair & Desk & Box & Board & Barrel & Stick & Chair & Desk & Box & Board & Barrel & Stick \\
    \hline
    Real & 5 & 6 & 9 & 5 & 9 & 4 & 157 & 213 & 200 & 128 & 206 & 58 \\
    \hline
    Synthetic & 418 & 408 & 376 & 589 & 602 & 596 & 1767 & 1344 & 1326 & 2123 & 1495 & 1961\\
    \hline
  \end{tabular}
  \caption{\textbf{Statistics on object in \dataset.}}
  \label{tab:obj}
\end{table*}

\begin{figure}[h]
  \centering
  \includegraphics[width=0.8\linewidth]{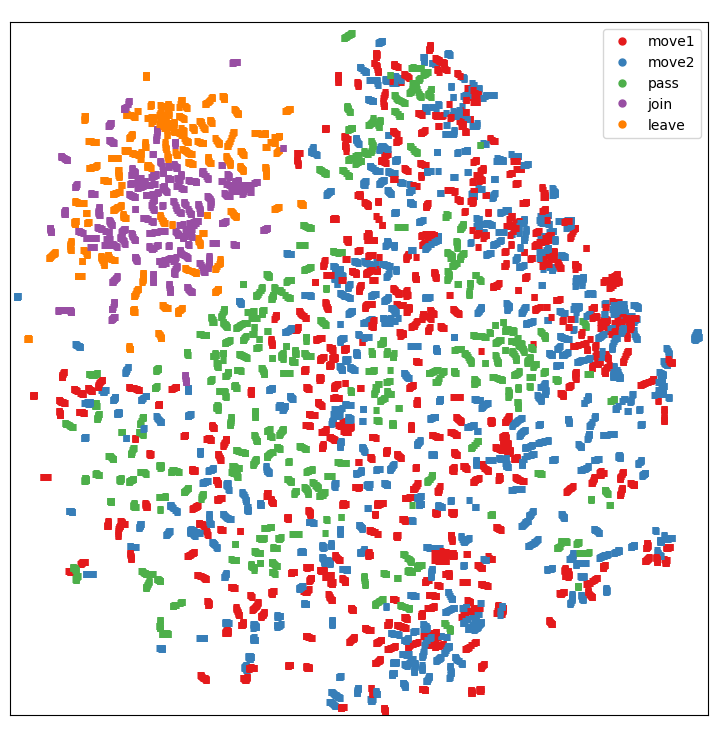}
  \caption{\textbf{T-SNE visualization of human poses for different collaboration modes.}}
  \label{fig:TSNE}
\end{figure}

\subsection{Collaboration Modes}

\dataset encompasses five human-human cooperation modes in collaborative object rearrangement. ``Move1'' refers to the scenario where two participants simultaneously rearrange objects and both are aware of the target. On the other hand, ``move2'' represents the scenario where objects are rearranged simultaneously, but only Person 1 knows the target. ``Pass'' indicates that one participant passes the object to another for relay transportation. ``Join'' means that Person 2 joins Person 1 in carrying the object during transportation. Lastly, ``leave'' signifies that Person 2 leaves during the joint transportation with Person 1.

According to the different durations of the two participants' contact with the object, ``move1'' and ``move2'' can be combined into collaborative carrying tasks. ``Pass'' represents the task of handover and solely moving the object.
Incorporating the join task and the leave task, \dataset totally comprises four different tasks (see Figure 4 in the main paper) based on the interaction between humans and objects. Fig. \ref{fig:vis} exemplifies the motions for each task.

As depicted in Fig. \ref{fig:TSNE}, distinct characteristics are exhibited by different cooperation modes in high-level movements, thereby offering an innovative standpoint and potential for comprehending and investigating collaborative behaviors.


\subsection{Participants}

31 participants, encompassing variations in height, weight, and gender, contributed to the capturing of \dataset-Real.

\subsection{Objects}

\dataset-Real has 38 objects while \dataset-Synthetic has about 3k objects. The objects encompass six categories, namely box, board, barrel, stick, chair, and desk, each exhibiting a rich diversity in surface shape and size. The distribution of object categories is detailed in Table \ref{tab:obj}. All the objects in \dataset-Real are shown in Fig. \ref{fig:obj_real}. Fig. \ref{fig:interpolation} shows samples from \dataset-Synthetic and their interpolation process.

\subsection{Camera Views}

\begin{figure}[h]
  \centering
  \includegraphics[width=\linewidth]{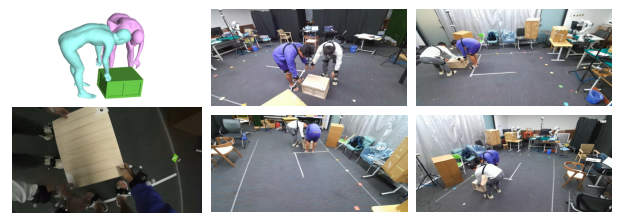}
  \caption{\textbf{Visualization of \dataset camera views.}}
  \label{fig:cam}
\end{figure}

\begin{figure}[h]
  \centering
  \includegraphics[width=\linewidth]{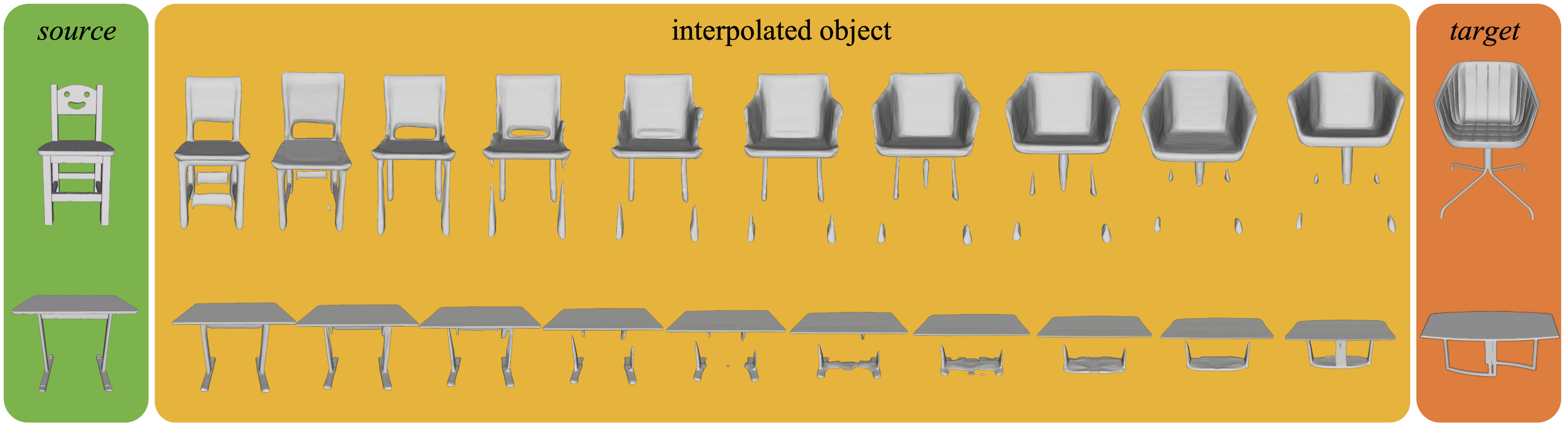}
  \caption{\textbf{Visualization of \dataset-Synthetic objects and interpolation.}}
  \label{fig:interpolation}
\end{figure}

\begin{figure}[h]
  \centering
  \includegraphics[width=\linewidth]{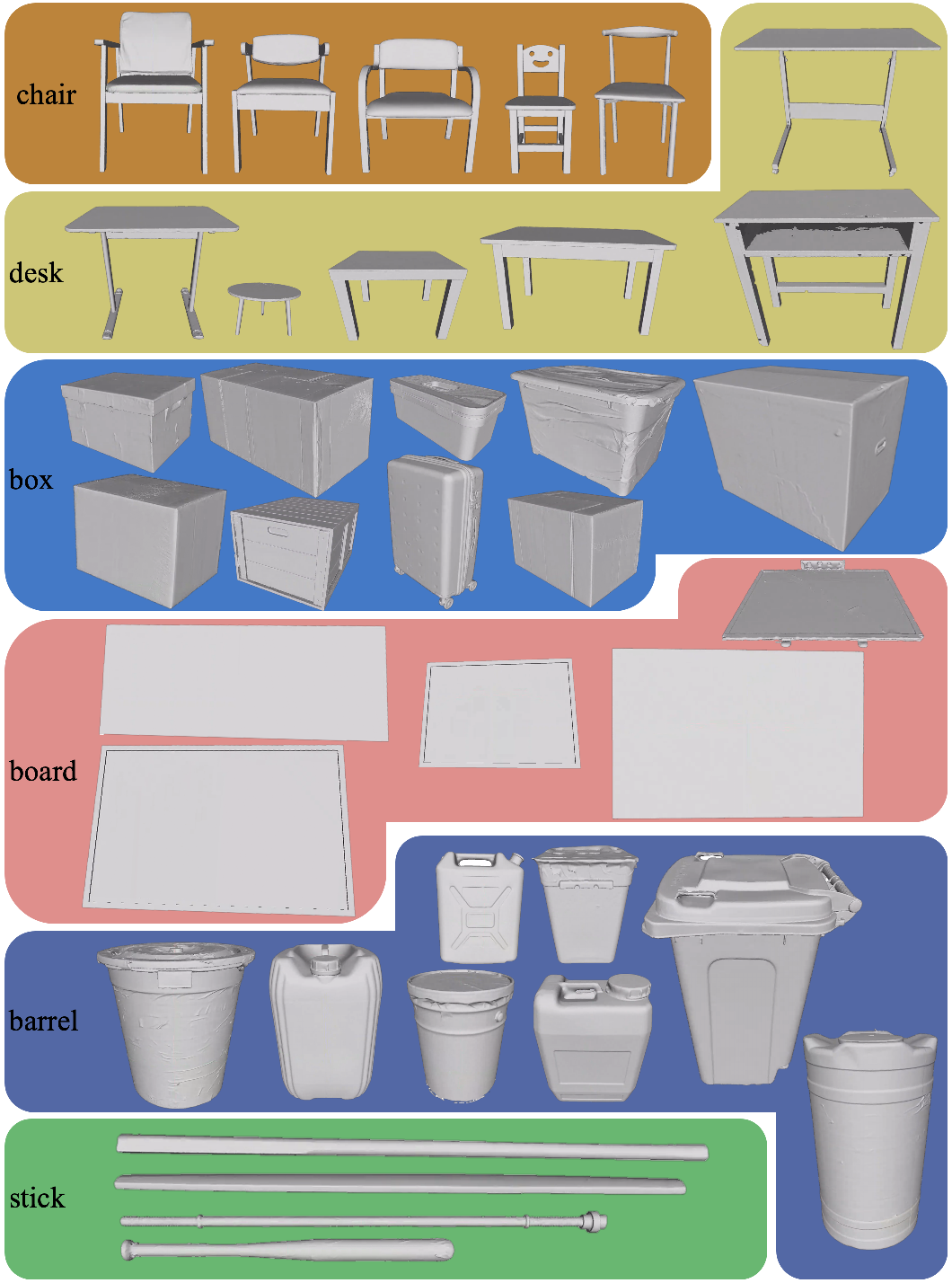}
  \caption{\textbf{Visualization of \dataset-Real objects.}}
  \label{fig:obj_real}
\end{figure}

\begin{figure}[h]
  \centering
  \includegraphics[width=\linewidth]{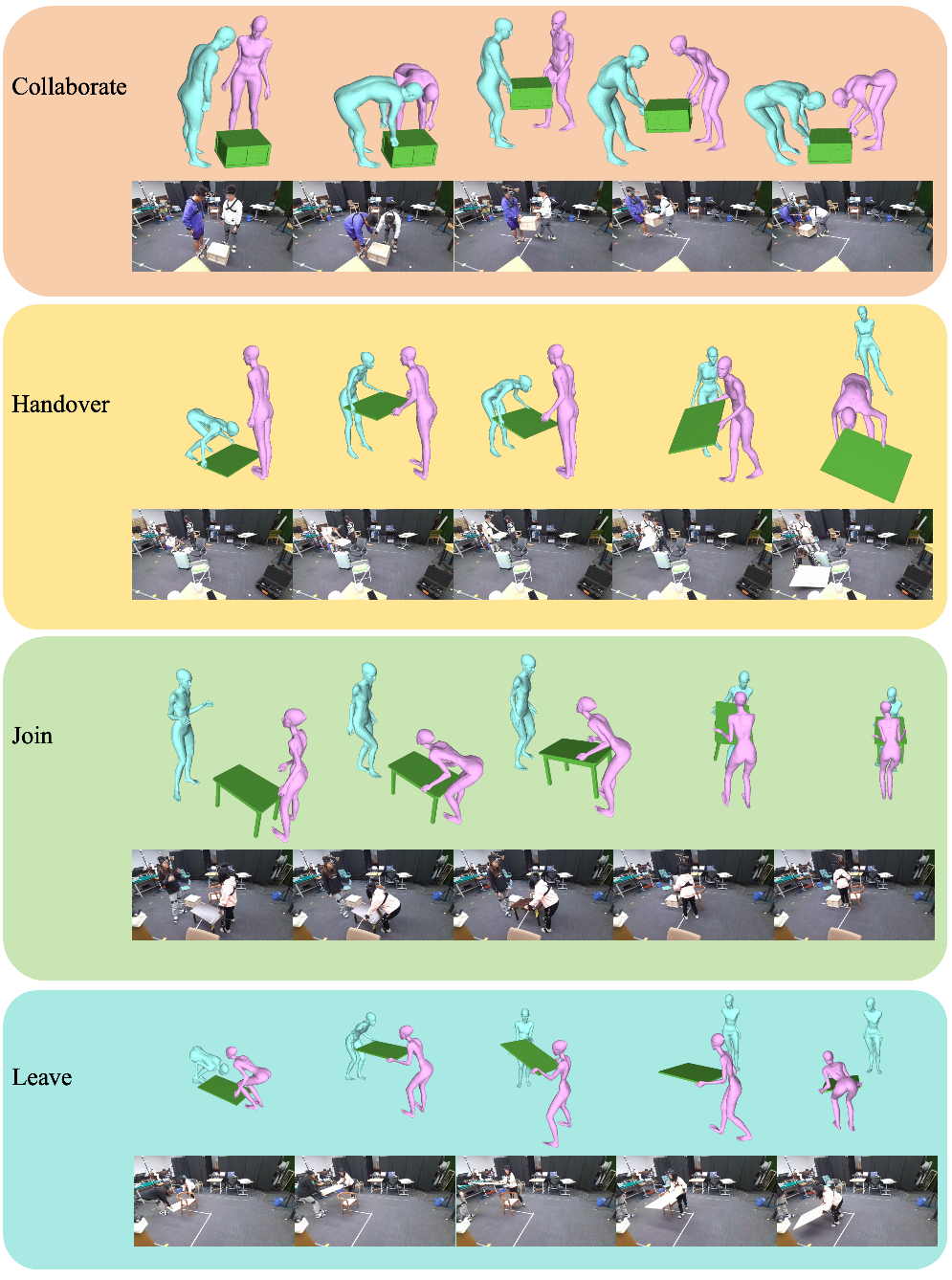}
  \caption{\textbf{Visualization of \dataset object rearrangement tasks.}}
  \label{fig:vis}
\end{figure}

Fig. \ref{fig:cam} shows the four allocentric and one egocentric views of our data capturing system.

\section{Details on Data Split}
\label{supp_sec:data_split_details}
Benefiting from the diverse temporal collaboration patterns from \dataset-Real and the large data amount of \dataset-Synthetic, we randomly select a subset of real object models and construct the training set as the combination of their real (T-Real) and synthesized (T-Synthetic) collaboration motion sequences. We formulate two test sets on \dataset-Real supporting studies of both non-generalization and inner-category generalization. The first test set (S1) consists of interaction performed on the objects that appear in the training set, while the second one (S2) is composed of interaction from novel objects. Detailed data distribution of each object category is shown in Table \ref{tab:train_test_split}.

\begin{table*}[tb]
  \centering
  \setlength{\tabcolsep}{4pt}
  \footnotesize
  \begin{tabular}{|c|cccccc|cccccc|}
    \hline
    \multirow{2}{*}{Set} & \multicolumn{6}{c|}{$\#$Object} & \multicolumn{6}{c}{$\#$Sequence} \\
    \cline{2-13}
    & Chair & Desk & Box & Board & Barrel & Stick & Chair & Desk & Box & Board & Barrel & Stick \\
    \hline
    T-Real & 3 & 4 & 6 & 3 & 6 & 2 & 93 & 104 & 96 & 51 & 113 & 25 \\
    \hline
    T-Synthetic & 418 & 408 & 376 & 589 & 602 & 596 & 1767 & 1344 & 1326 & 2123 & 1495 & 1961\\
    \hline
    S1 & 3 & 4 & 6 & 3 & 6 & 2 & 40 & 62 & 45 & 21 & 51 & 6 \\
    \hline
    S2 & 2 & 2 & 3 & 2 & 3 & 2 & 24 & 47 & 59 & 56 & 42 & 27 \\
    \hline
  \end{tabular}
  \caption{\textbf{Train-test split on \dataset.}}
  \label{tab:train_test_split}
\end{table*}

\section{Evaluation Metrics for Benchmarks}
\label{supp_sec:metrics_benchmarks}

\subsection{Human-object Motion Forecasting}
\label{supp_sec:evaluation_metrics_motion_forecasting}

Evaluation metrics include the human joints position error $J_e$, the object translation error $T_e$, the object rotation error $R_e$, the human-object contact accuracy $C_{\text{acc}}$, and the penetration rate $P_r$.

\begin{itemize}

\item We define $J_e$ as the average Mean Per Joint Position Error (MPJPE) of the two persons. MPJPE represents the mean per-joint position error of the predicted human joint positions and the ground-truth values. The unit of $J_e$ is one millimeter.

\item Translation error ($T_e$) and rotation error ($R_e$) denote the average L2 difference between the predicted object translation vectors and the ground-truth ones, and the average geodesic difference between the estimated object rotation matrices and the ground-truth ones, respectively. The unit of $T_e$ is one millimeter. The unit of $R_e$ is one degree.

\item Physical metrics: To assess contact fidelity, we detect contacts on the two hands of the two persons for each frame with an empirically designed distance threshold (5 centimeters). We then examine the contact accuracy ($C_{acc}$), which indicates the average percentage of contact detection errors in the predicted motions. Additionally, we examine the object penetration ratio ($P_r$) representing the mean percentage of object vertices inside the human meshes. The units of the two metrics are percentages.

\end{itemize}

\subsection{Interaction Synthesis}
\label{supp_sec:evaluation_metrics_interaction_synthesis}

Following an existing individual human-object interaction synthesis study~\cite{OMOMO}, the evaluation metrics include the root-relative human joint position error $RR.J_e$, the root-relative human vertex position error $RR.V_e$, the human-object contact accuracy $C_{\text{acc}}$, and the FID score (\textit{FID}).

\begin{itemize}

\item $RR.J_e$ denotes the average root-relative MPJPE of the two persons. The root-relative MPJPE represents the mean per-joint position error of the predicted human joint positions relative to the human root position and the ground-truth values. The unit of $RR.J_e$ is one millimeter.

\item $RR.V_e$ denotes the average root-relative Mean Per Vertex Position Error (MPVPE) of the two persons. The root-relative MPVPE represents the mean per-vertex position error of the predicted human vertex positions relative to the human root position and the ground-truth values. The unit of $RR.V_e$ is one millimeter.

\item $C_{\text{acc}}$ is the same as that in Section~\ref{supp_sec:evaluation_metrics_motion_forecasting}.

\item The Fr{\'e}chet Inception Distance (\textit{FID}) quantitatively evaluates the naturalness of synthesized human motions. We first train a feature extractor on \dataset-Real to encode each human-object-human motion sequence to a 256D feature vector $\bar{f}_i$ and acquire the ground-truth human motion feature distribution $\bar{D}$=$\{\bar{f}_i\}$. We then replace the motions of the two persons as synthesized ones and obtain another distribution $D$=$\{f_i\}$. Eventually, the \textit{FID} denotes the 2-Wasserstein distance between $\bar{D}$ and $D$. Since \dataset-Real provides action labels, the feature extractor is supervised-trained by fulfilling the action recognition task. The network structure of the feature extractor is a single-layer Transformer~\cite{vaswani2017attention}.

\end{itemize}

\section{Qualitative Results on Benchmarks}
\label{supp_sec:qualitative_results_on_benchmarks}

\begin{figure*}[h]
  \centering
  \includegraphics[width=0.7\textwidth]{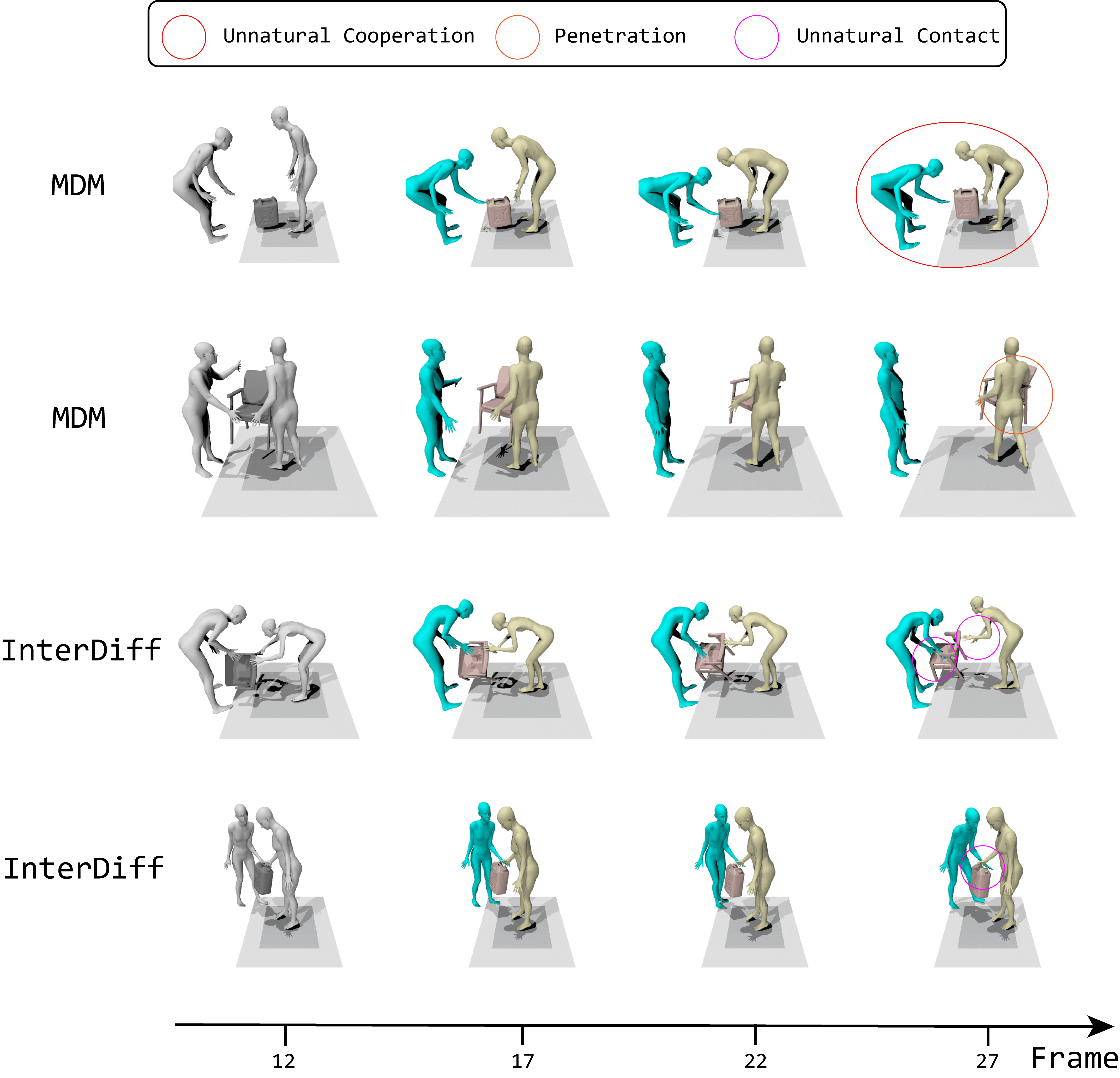}
  \caption{\textbf{Qualitative results of human-object motion forecasting.} Grey meshes are from the task inputs.}
  \label{fig:vis_forecasting}
\end{figure*}

\begin{figure*}[h]
  \centering
  \includegraphics[width=\textwidth]{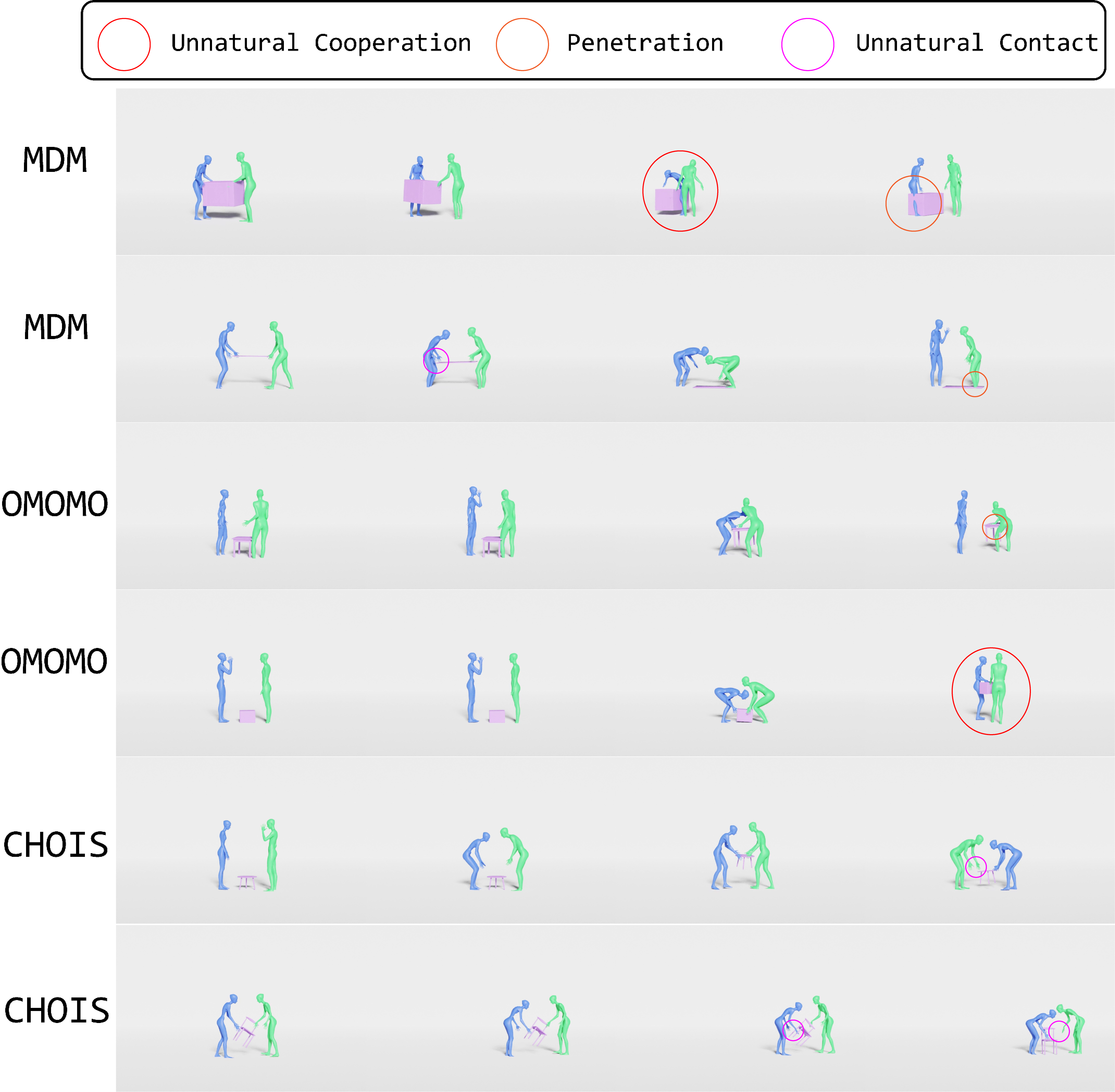}
  \caption{\textbf{Qualitative results of interaction synthesis.}}
  \label{fig:vis_synthesis}
\end{figure*}

Figure \ref{fig:vis_forecasting} and Figure \ref{fig:vis_synthesis} exemplify generated motions for the human-object motion forecasting task and the interaction synthesis task, respectively. Since the baseline methods do not focus on generating hand poses, we replace hand poses in ground truth with flat hands to facilitate fair comparisons. Despite diverse cooperation modes that can be generated, the baseline methods consistently encompass unsatisfactory performances including unnatural collaboration, inter-penetration, and unnatural contact.

\section{Details on the Application of \dataset-Synthetic}
\label{supp_sec:application_details}



To evaluate the application of \dataset-Synthetic, we use the lightweight CAHMP~\cite{corona2020contextaware} to conduct the motion forecasting experiments. 
Unlike the experiments in section \textbf{Human-object Motion Forecasting} mentioned in the main paper, where 15 frames are predicted, here we predict the human-object motion for the next 10 frames given the previous 10 frames.

\subsection{Task Formulation} 
Given the object's 3D model and human-object poses in adjacent 10 frames, the task is to predict their subsequent poses in the following 10 frames. The human pose $P_h \in \mathbb{R}^{23 \times 3}$ represents the joint rotations of the SMPL-X~\cite{SMPLX} model, while the object pose $P_o = \{R_o \in \mathbb{R}^3, T_o \in \mathbb{R}^3\}$ denotes 3D orientation and 3D translation of the rigid object model.

\subsection{Evaluation Metrics} 
Following existing motion forecasting works~\cite{CAHMP,HO-GCN,InterDiff}, we evaluate human joints position error $J_e$, object translation error $T_e$, object rotation error $R_e$. Details of the three metrics can be found in Section~\ref{supp_sec:evaluation_metrics_motion_forecasting}.

\subsection{Results}
Comparing the 1K real dataset with the 0.1K real dataset supplemented with synthetic data generated through retargeting, we observed that the quality of the synthetic data is comparable to the real data. Additionally, due to the increased diversity of objects and enriched spatial relations between humans and objects in the synthetic data, it exhibits better generalization performance in object motion forecasting.

Comparing the evaluation results of the 1K real dataset with the results obtained by augmenting it with additional 4K synthetic data, we observed a significant performance gain from the synthetic data. This demonstrates that the inclusion of synthetic data enhances the value of our dataset and better supports downstream tasks.

\section{Details on Humanoid Skill Learning using \dataset}
\label{supp_sec:details_on_humanoid_skill_learning}

As introduced in Section 5.2 in the main paper, we use \dataset's human-box interaction data to facilitate humanoid skill learning for box lifting. This section presents details on this application, including the simulation environment configuration (Section~\ref{supp_sec:humanoid_environment_setup}), task formulation and evaluation (Section~\ref{supp_sec:humanoid_task_formulation_and_evaluation}), adapting human interaction data to humanoid (Section~\ref{supp_sec:humanoid_retargeting},~\ref{supp_sec:humanoid_tracking}), benchmark method designs (Section~\ref{supp_sec:humanoid_IL}), and experiments (Section~\ref{supp_sec:humanoid_experiments}).

\subsection{Environment Setup}
\label{supp_sec:humanoid_environment_setup}

We use the popular Unitree H1 humanoid robot~\cite{unitree_h1} in Isaac Gym~\cite{makoviychuk2021isaac} simulation environment. The H1 humanoid has 19 revolute joints with fixed limits on motion ranges. The interaction scene contains the humanoid, a box weighing 0.5kg initially posed on the floor, and a third-person-view camera providing visual signals for the skill policy.

\textbf{Data clipping:} The original interactions in CORE4D are collaborations of two persons. In this application, we regard each of them as two individual human-object interaction motions and obtain the box-lifting clips automatically by measuring hand-object distances and the object's height. We discard the motion if the individual is not touching the object during its lifting process.

\textbf{Train-test split:} As a result, we acquire 890 individual human-box lifting data clips covering 22 boxes augmented from three real-world ones. We select 16 boxes (504 clips) as the training set and the remains as the test set. The training set is used only to train the skill policy (Section~\ref{supp_sec:humanoid_IL}) and its preceding motion tracker (Section~\ref{supp_sec:humanoid_tracking}), while the test set is used only to evaluate the skill policy to assess its generalizability to unseen box shapes.

\textbf{Observations:} The real-time input of the humanoid skill policy is $\mathcal{I} = \{s_{\text{proprio}}, r_{\text{root}}, t_{\text{root}}, \mathcal{C}, \mathcal{D}\}$, where $s_{\text{proprio}} \in \mathbb{R}^{19\times2}$ is humanoid's proprioception, $r_{\text{root}} \in \mathbb{R}^{3}$ and $t_{\text{root}} \in \mathbb{R}^{3}$ are humanoid's root orientation and position in the world coordinate system respectively, and $\mathcal{C} \in \mathbb{R}^{360 \times 480 \times 3}$ and $\mathcal{D} \in \mathbb{R}^{360 \times 480}$ are the color and depth image captured by the camera. Specifically, $s_{\text{proprio}}$ is composed of the angles and velocities of each joint.

\textbf{Action:} Given the real-time input $\mathcal{I}$, the skill policy needs to generate an action $\mathcal{A}$ and use it to actuate the humanoid in the simulation environment. The action $\mathcal{A} \in \mathbb{R}^{19}$ is defined as 19 DoF joint angles, which are transferred to joint torques via a PD controller with pre-defined proportional-derivative gains~\cite{fu2024humanplus}. The simulation environment uses the computed joint torques to actuate the humanoid.

\subsection{Task Formulation and Evaluation Metric}
\label{supp_sec:humanoid_task_formulation_and_evaluation}

Given an unseen box in the scene and a starting position, the humanoid is required to adjust its pose, touch the box, and finally lift it larger than 20 centimeters. The evaluation metric (\textit{SR}) is the task success rate defined as whether the box reaches 20 centimeters higher than its initial position.

\subsection{Retargeting Interactions from \dataset onto H1 Humanoid Robot}
\label{supp_sec:humanoid_retargeting}

Following existing humanoid skill learning advances~\cite{he2024omnih2o,he2024learning,tang2024humanmimic}, we use an optimization strategy to solve the retargeting problem. The optimization is finding the optimal sequence of $\{s_{\text{proprio}}, r_{\text{root}}, t_{\text{root}}\}$ that can minimize differences between motions of the human $A$ and the humanoid $B$ on positions of $K$ paired joints in the world coordinate system $L_p = \sum\limits_{k=1}^{K} \Vert P_{a_k} - P_{b_k} \Vert^2$, where $P \in \mathbb{R}^{T\times 3} $ denotes the position sequence of a joint, and $T$ is the frame number. $<a_k, b_k>$ is a pre-defined pair of joints with similar semantics, where $a_k$ is from the human skeleton, and $b_k$ is from the humanoid skeleton.

We design a temporal loss $L_t = \sum\limits_{t=2}^{T} \sum\limits_{i=1}^{19} (\Vert s_{\text{proprio},t,i} - s_{\text{proprio},t-1,i} \Vert^2 + \Vert r_{\text{root},t,i} - r_{\text{root},t-1,i} \Vert^2 + \Vert t_{\text{root},t,i} - t_{\text{root},t-1,i} \Vert^2)$ to further improve motion smoothness. The overall optimization target is $L_p + L_t$.

As a result, this retargeting method converts human-box interaction motions to humanoid-box interaction animations. The animations are not physically realistic, and we use the upcoming tracking method to obtain physically realistic ones following HumanPlus~\cite{fu2024humanplus}.

\subsection{Tracking Humanoid-box Interaction Animations}
\label{supp_sec:humanoid_tracking}

Given a humanoid-box interaction animation, the motion tracking methods~\cite{peng2018deepmimic,phc,tessler2024maskedmimic} control the humanoid in the simulation environment and generate the motion that mostly closely resembles the animation. The generated motions are both physically realistic and able to fulfill the task via mimicking animations, which are utilized as skill demonstrations in the following skill policy learning.

We select HST~\cite{fu2024humanplus} as our tracker. The official implementation of the HST~\cite{fu2024humanplus} tracker fails to lift the box, and we design an improved HST that can successfully track the animations and control the humanoid to lift the box physically. We describe the key designs of our tracker below.

\textbf{Input:} In each frame, the tracker inputs the states of the real-time humanoid $S_h$, the humanoid animation target $\hat{S_h}$, the real-time box $S_b$, and the box target $\hat{S_b}$. $S_h=\{r_{\text{root}},\dot{r}_{\text{root}}, t_{\text{root}}, \dot{t}_{\text{root}},J,\dot{J},R,\dot{R},T,\dot{T}\}$, where $r_{\text{root}} \in \mathbb{R}^3$, $\dot{r}_{\text{root}} \in \mathbb{R}^3$, $t_{\text{root}} \in \mathbb{R}^3$, $\dot{t}_{\text{root}} \in \mathbb{R}^3$, $J \in \mathbb{R}^{19}$, $\dot{J} \in \mathbb{R}^{19}$, $R \in \mathbb{R}^{(19+1)\times3}$, $\dot{R} \in \mathbb{R}^{(19+1)\times3}$, $T \in \mathbb{R}^{(19+1)\times3}$, $\dot{T} \in \mathbb{R}^{(19+1)\times3}$ denotes root's global orientation, root's global angular velocity, root's global position, root's global linear velocity, joint angle, joint velocity, global positions of the root and joints, global linear velocities of the root and joints, global orientations of the root and joints, and global angular velocities of the root and joints. $\hat{S_h}$ is formulated the same as $S_h$. $S_b=\{r_b,\hat{r}_b,t_b,\hat{t}_b\}$, where $r_b$, $\hat{r}_b$, $t_b$, $\hat{t}_b$ denotes box global orientation, box global angular velocity, box global position, and box global linear velocity, respectively. $\hat{S_b}$ shares the same definition with $S_b$.

\textbf{Output:} In each frame, the tracker outputs a 19 DoF action vector representing target joint angles. The joint torques are computed via a low-level PD controller and are fed into the simulation environment to actuate the humanoid.

\textbf{Reward function:} The tracker is trained via a reinforcement learning (RL) method PPO~\cite{rl_ppo}. The reward design is the most crucial part of the method's performance. Using original rewards from HumanPlus~\cite{fu2024humanplus} cannot lift the box successfully due to inaccurate tracked hand positions. To handle this issue, we draw inspiration from PhysHOI~\cite{wang2023physhoi} and use a multiplication of humanoid reward $\mathcal{R}_h$, box reward $\mathcal{R}_b$, and humanoid-box interaction reward $\mathcal{R}_i$ as the overall reward $\mathcal{R}_{\text{overall}}$: $\mathcal{R}_{\text{overall}} = \mathcal{R}_h \times \mathcal{R}_b \times \mathcal{R}_i$.

\begin{itemize}
\item $\mathcal{R}_h = 0.5\exp(-5\Vert T-\hat{T} \Vert_1) +5\exp(-10\Vert t_{\text{root}} - \hat{t}_{\text{root}} \Vert_1) +10\exp(-10\Vert J_l - \hat{J}_l\Vert_1) +10\exp(-10\Vert J_r - \hat{J}_r\Vert_1)$, where $l$/$r$ denotes the left/right hand.
\item $\mathcal{R}_b = \exp(-10\Vert t_b - \hat{t}_b \Vert_1)$.
\item $\mathcal{R}_i = \exp(-10\Vert P_{l,h\rightarrow b}-\hat{P}_{l,h\rightarrow b} \Vert_1 - 10\Vert P_{r,h\rightarrow b}-\hat{P}_{r,h\rightarrow b} \Vert_1)$, where $P_{h\rightarrow b}$ and $\hat{P}_{h\rightarrow b}$ denotes the humanoid joint position in the box's coordinate system, and $l$/$r$ denotes the left/right hand.
\end{itemize}

\textbf{Training strategies:} We adopt an early-termination strategy from DeepMimic~\cite{peng2018deepmimic} that terminates the rollout when the humanoid is 0.5 meters away from its target or its root height is below 0.5 meters.

\subsection{Reinforcement Learning and Imitation Learning Method Designs}
\label{supp_sec:humanoid_IL}

Combining the retargeting method (Section~\ref{supp_sec:humanoid_retargeting}) with the improved HST tracker (Section~\ref{supp_sec:humanoid_tracking}), we transfer CORE4D's data to physically realistic humanoid box-lifting demonstrations. The final step is to train a skill policy that mimics the demonstrations and can lift unseen boxes in test time. We select two \textbf{vision-based} imitation learning (IL) methods, HIT~\cite{fu2024humanplus} and ACT~\cite{ACT}, and use their official implementations.

To examine the value of the demonstrations, we compare the two IL methods with a commonly used \textbf{state-based} RL algorithm PPO~\cite{rl_ppo}. The PPO is implemented with the code from HumanPlus~\cite{fu2024humanplus}, with a change on the reward design $\mathcal{R}$: $\mathcal{R} = \mathcal{R}_b + \mathcal{R}_{\textbf{success}}+\mathcal{R}_{i}+\mathcal{R_\textbf{alive}}$, where:

\begin{itemize}
\item $\mathcal{R}_b = \exp(\Vert t_b - \hat{t}_b \Vert_2^2)$, where $\hat{t}_b$ is the pre-defined target box center position for the task.
\item $\mathcal{R}_{\textbf{success}} = [\mathcal{R}_d < 0.01]$ encouraging achieving the task.
\item $\mathcal{R}_i = -0.1(\Vert P_{l,h\rightarrow b} \Vert_2^2 +\Vert P_{r,h\rightarrow b} \Vert_2^2)$, where $P_{h\rightarrow b}$ and $\hat{P}_{h\rightarrow b}$ denotes the humanoid joint position in the box's coordinate system, and $l$/$r$ denotes the left/right hand. This reward encourages humanoid hands to explore near the object.
\item $\mathcal{R_\textbf{alive}} = 0.1$ encouraging the humanoid being alive.
\end{itemize}

\subsection{Experiments}
\label{supp_sec:humanoid_experiments}

The evaluation results are shown in Table 6 and Figure 6 in the main paper. Leveraging \dataset data, the policy can achieve 21.0$\%$ (for HIT) and 26.5$\%$ (for ACT) task success rates, which are significantly larger than that of data-free PPO (0.0$\%$), demonstrating the value of \dataset for humanoid box-lifting skill learning.

\section{\dataset-Real Data Capturing Instructions}
\label{supp_sec:data_capture_instructions_and_costs}

\noindent\textbf{Target.} We divide a $4m \times 5m$ field into 20 squares and number them, and place colored labels as markers along the perimeter of the field. The following language instructs participants: \textit{"Please collaboratively move the object to the target square. You can choose any path and orientation of the object as you like. It is not necessary to be overly precise with the final position - a rough placement is fine. Do not make unnatural motions just to achieve an exact position. Do not use verbal communication with each other."}. As for the settings when only one participant knows the target, the target square number is written on a piece of paper and shown to the participant who knows the target. And additional instructions are given as: \textit{"If you know the target, do not use language or direct body language to inform the other party (such as pointing out the location). If you do not know the target, please assist the other participant in completing the transportation."}.

\noindent\textbf{Collaboration Mode.} The instructions are given as follows to indicate different Collaboration Modes for the participants. For Collaborate mode: \textit{"Based on the target, please cooperatively transport the object, or upright any overturned tables, chairs, etc. Both participants should be in contact with the object throughout the process."}. For Handover mode: \textit{"Please decide the handover point yourselves, then have one person hand the object to the other, completing the object transfer in relay."}. For Leave and Join modes: \textit{"One person will transport the object throughout, while the other leaves or joins to help at a time point not disclosed to the collaborator."}.

\noindent\textbf{Obstacle.} The instructions are given as follows to guide the participants in tackling obstacles: \textit{"There are a varying number of obstacles on the field. If they get in your way, please decide on your own how to solve it using some common everyday operations. If the obstacles occupy the destination, please place the object near the destination."}.

\end{document}